\documentclass[10pt,letterpaper]{article}
\PassOptionsToPackage{table}{xcolor}
\usepackage{kps-techreport}
\usepackage{float}
\title{KUPAS MASTER: Distilling the Tacit Expertise of Master Practitioners into Agent-Ready Experience Corpora}
\author{KUPAS MASTER Team\footnotemark[1]}
\affiliation{Shanghai Kupas Technology Co., Ltd. and Tongji University}
\reportinfo{Technical Report\footnotemark[2], September 2026}
\leftlogo{kps_logo}
\rightlogos{\includegraphics[height=\rightlogoheight]{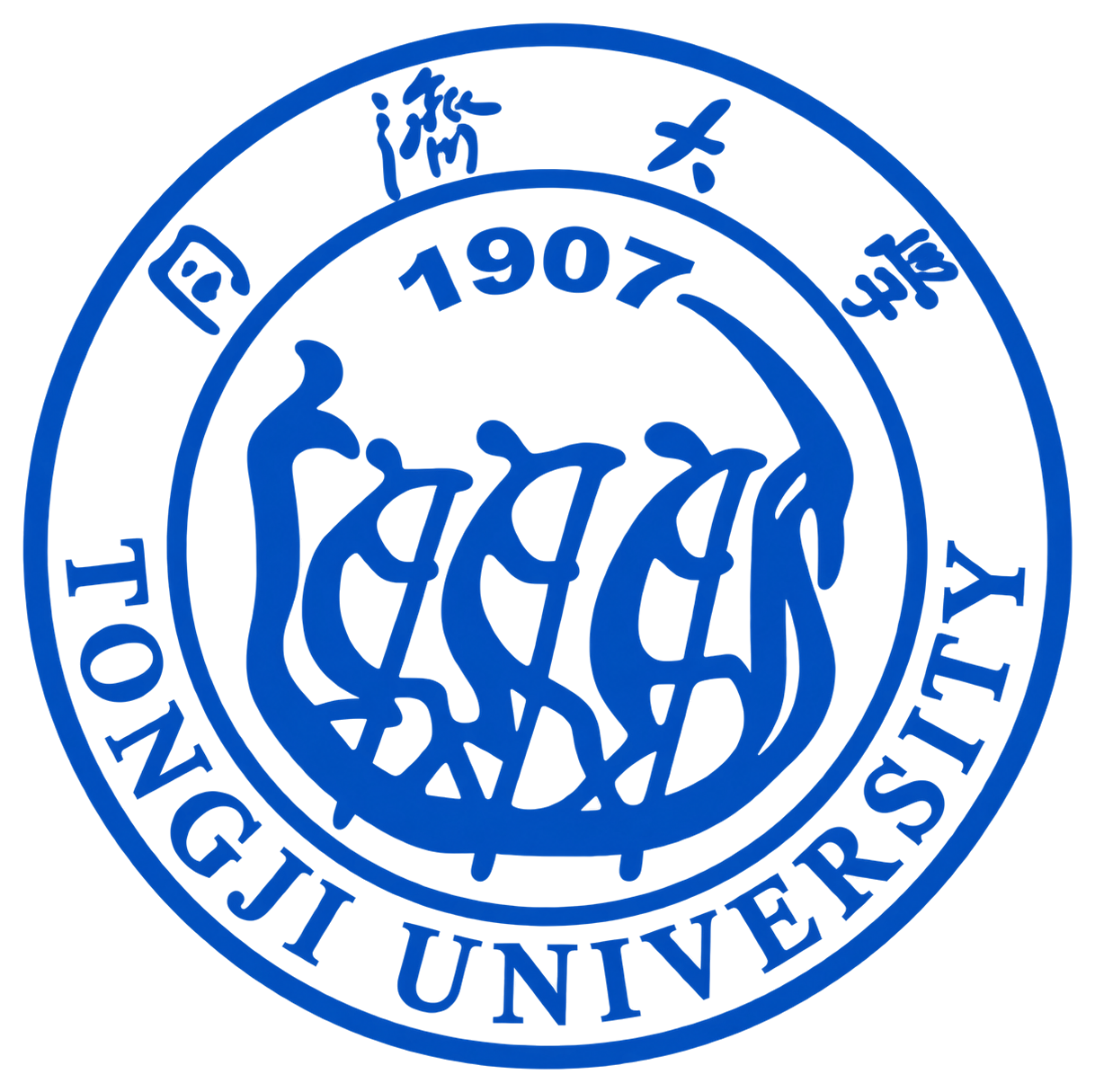}\hspace{10pt}\includegraphics[height=\rightlogoheight]{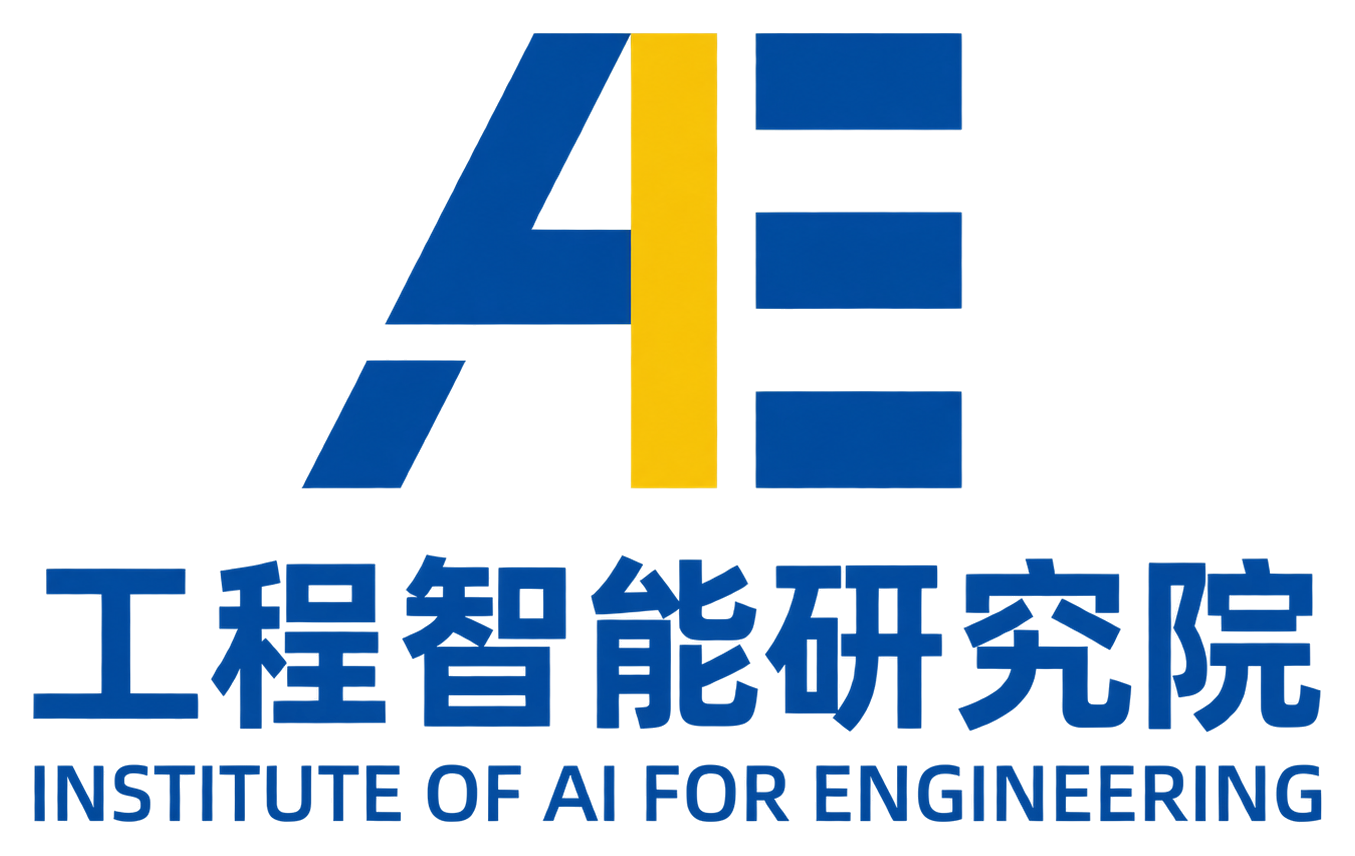}}

\newcommand{\product}{KUPAS MASTER\xspace}

\definecolor{experimentred}{HTML}{C31C28}
\definecolor{kpstodo}{HTML}{C31C28}
\definecolor{kpstodobg}{HTML}{FFF3F3}

\newcommand{\asset}{\mathcal{A}}

\newcommand{\reportfigure}[3]{\begin{figure}[htbp]\includegraphics[width=\linewidth]{figs/#1.pdf}\caption{#2}\label{#3}\end{figure}}
\hypersetup{pdftitle={KUPAS MASTER: Distilling the Tacit Expertise of Master Practitioners into Agent-Ready Experience Corpora},pdfauthor={KUPAS MASTER Team},pdfsubject={Commercial technical report on expert experience engineering}}

\ifPDFTeX\pdfminorversion=7\else
  \AtBeginDocument{}
\fi

\ifXeTeX
  \usepackage{xeCJK}
  \setCJKmonofont{KUPAS_MASTER_Chinese-Regular.ttf}[Path=fonts/]
\else
  \PackageError{KUPAS MASTER}{Use XeLaTeX or Tectonic for this multilingual report}
    {The bibliography now contains Chinese source titles.}
\fi

\ifPDFTeX
  \newcommand{\lsftablelatin}{\fontfamily{cmr}\selectfont}
\else
  \newfontfamily\lsftablelatin{lmroman10-regular.otf}[
    BoldFont=lmroman10-bold.otf]
\fi
\newcommand{\lsftableheading}{\sffamily}
\newcommand{\lsftablehead}[1]{{\lsftableheading\bfseries #1}}
\definecolor{lsftableshade}{HTML}{F4F4F4}
\newcommand{\lsftablefont}{\lsftablelatin\fontsize{9}{11}\selectfont}
\newcommand{\lsftablesetup}{%
  \lsftablefont\color{black}%
  \renewcommand{\arraystretch}{1.08}%
  \setlength{\tabcolsep}{5pt}%
  \setlength{\extrarowheight}{0pt}%
  \setlength{\heavyrulewidth}{0.8pt}%
  \setlength{\lightrulewidth}{0.5pt}%
  \setlength{\cmidrulewidth}{0.3pt}%
  \setlength{\aboverulesep}{0.4ex}%
  \setlength{\belowrulesep}{0.65ex}%
  \arrayrulecolor{black}}
\DeclareCaptionFont{lsftable}{\lsftablefont}
\DeclareCaptionFont{lsftablelabel}{\lsftableheading\bfseries}
\AtBeginEnvironment{table}{\lsftablesetup}
\AtBeginEnvironment{table*}{\lsftablesetup}
\AtBeginEnvironment{tabularx}{%
  \lsftablesetup\rowcolors{2}{white}{lsftableshade}}
\AtBeginEnvironment{longtable}{%
  \lsftablesetup\rowcolors{3}{lsftableshade}{white}}

\begin{document}
\maketitle
\footnotetext[1]{Detailed team membership is listed in Appendix~\ref{app:team}.}
\footnotetext[2]{Official website: \url{https://lsf.kupasai.com/}. Report homepage: \url{https://tongjiai4e.github.io/KUPAS-MASTER-Report/}.}
\setcounter{footnote}{2}
\begin{abstractbox}
In every field, experienced professionals know more than just facts and conclusions. They know which cues matter, why a judgment is reasonable, which action to take, and when a familiar approach no longer applies. Routine work records often leave out this tacit knowledge, making it difficult for large language model (LLM) agents to use professional experience effectively. We introduce \product, an experience engineering platform built around nine-layer cognitive corpus construction. It turns heterogeneous work records and practitioner interviews into traceable, reusable experience corpora for agents. Six case elements preserve the task process: context, cues, judgment, action, boundaries, and outcomes. Nine-layer cognitive corpus construction organizes tacit experience along nine extraction dimensions and stores the resulting assets in six libraries: rules, constraints, best practices, negative examples, corner cases, and skills. Semantic alignment, individual experience distillation, organizational consolidation, and cross-review preserve source evidence, conditions of use, and unresolved disagreements. The platform packages these assets into callable skills with explicit inputs, steps, dependencies, and stopping conditions, connecting experience collection to task execution and evaluation feedback. Using authorized samples from 20 randomly selected practitioners, the platform processed 1,576 source files into 23,024 individual experience records and 13,113 organizational assets. The evaluation spans 177 questions and 531 responses across multiple professional domains. Under common task inputs and scoring criteria, the base model, raw-corpus retrieval-augmented generation (RAG), and \product agent scored 70.63, 79.75, and 89.58, respectively. The \product agent improved on raw-corpus RAG by 9.83 points, with gains in all seven scoring dimensions. The systematic comparison demonstrates the effectiveness of the platform and its core nine-layer method in professional tasks, delivering better task quality, deeper professional judgment, and effective experience reuse. The platform provides a practical path from individual tacit experience to organizational knowledge and agent capabilities.
\end{abstractbox}

\section{Introduction}
\label{sec:introduction}
As agents enter enterprise and scientific workflows, tasks that depend on professional experience require them to understand context, make sound judgments, and choose appropriate actions. Much of the experience they need is spread across practitioners and work records. Turning it into traceable, reusable corpora strengthens task execution and preserves critical know-how. Major national initiatives reflect the international importance of this need. China's AI Plus initiative calls for reusable expert knowledge and high-quality datasets~\citep{state2025aiplus}. In the United States, the Genesis Mission calls for an integrated AI platform that uses federal scientific datasets to develop scientific foundation models and agents for hypothesis testing and research automation~\citep{whitehouse2025genesis}. Together, these initiatives highlight a shared priority: making domain knowledge usable by AI systems. By turning professional experience into reusable assets for agents, \product addresses a problem of global significance for industrial productivity and scientific innovation.

Organizing professional experience for AI systems is therefore an important problem in enterprise knowledge management. A maintenance expert may combine an unusual sound, recent repair records, and operating load to diagnose a fault. A mediator may first verify disputed facts before discussing responsibility. In both tasks, useful experience includes the evidence behind a decision, the alternatives considered, the conditions for an action, and the reasons to stop or change course. Without those conditions, another practitioner or agent may apply a useful rule to the wrong case.

Organizations already keep manuals, work orders, incident reports, recordings, and case reviews. Yet terminology varies, intermediate decisions go unrecorded, and observations made at the time may be mixed with later explanations. A record of a successful intervention may omit the conditions that made it work. Rare but serious failures may survive only in conversation and remain unavailable to text retrieval. Knowledge acquisition research has long studied how to recover such information from actual work. The Critical Decision Method, for example, uses structured questions about specific incidents to identify the cues and reasoning that experienced practitioners relied on~\citep{klein1989critical,hoffman1998cdm,militello1998acta}.

Large language models (LLMs) offer new ways to organize and use these materials. Retrieval-augmented generation (RAG) brings external text into inference~\citep{lewis2020rag}, graph retrieval uses relationships across records~\citep{edge2024graphrag}, and agent frameworks connect reasoning with tool use~\citep{yao2023react,gao2025agentscope}. These methods help retrieve and use existing information. Several questions remain about the experience itself: how to fill gaps, express conditions of use, handle conflicting accounts, and check whether an extracted procedure can actually be followed. Answering them requires attention to how the underlying experience was collected, organized, and reviewed.

\product calls this process \emph{experience engineering}: acquiring, structuring, validating, and maintaining experience so that it becomes reusable, reviewable knowledge. Here, an experienced practitioner is someone who can contribute practical knowledge about a specific task and context. The platform links that knowledge to its task, conditions, and supporting evidence so that others can understand and review the judgment before reusing it. It also records additions and revisions through version control, supporting expert review, source tracing, and later agent use.

For this report, we randomly selected 20 authorized platform users from different professional domains and examined their corpora, assets, and run records. The sample contains 1,576 source files, 30,762 corpus chunks, 23,024 individual experience records, and 13,113 organizational assets. These counts correspond to source collection, parsing, experience extraction, and organizational consolidation.

We compare three agent configurations on the same tasks under a shared scoring rubric: a base-model agent (A), an agent using raw-corpus RAG (B), and an agent using \product experience assets and skills (C). The evaluation contains 177 questions and 531 responses. We average the platform-reported practitioner scores equally across the 20 practitioners; each score combines seven weighted dimensions. The scores are 70.63, 79.75, and 89.58 for A, B, and C. The results demonstrate the effectiveness of nine-layer cognitive corpus construction, with consistent improvements in task quality, professional judgment, and execution. Later sections explain the evaluation and the role of experience assets in these gains.

The platform makes four main contributions. \textbf{A structured representation of expert experience} connects six case elements, nine extraction dimensions, and six asset types in a common format for collection, review, tracing, and execution. \textbf{An experience corpus construction workflow} converts heterogeneous work records into candidate assets while retaining their evidence, context, and review status. \textbf{Organizational consolidation} compares and combines experience under explicit task conditions. It keeps context-dependent alternatives separate and flags conflicts, disagreements, and insufficient evidence for review. Finally, \textbf{skill construction and evaluation} turns these assets into callable procedures and compares their use against the base model and raw-corpus RAG.

The following sections introduce task scope, experience representation, corpus construction, and organizational consolidation, with examples from the authorized sample. Section~\ref{sec:evaluation} describes the evaluation design, including controls for model configuration, input information, and runtime resources.

\section{System Overview and Task Scope}
\label{sec:overview}
\subsection{Using the system}
\begin{figure}[!t]
\includegraphics[width=\linewidth]{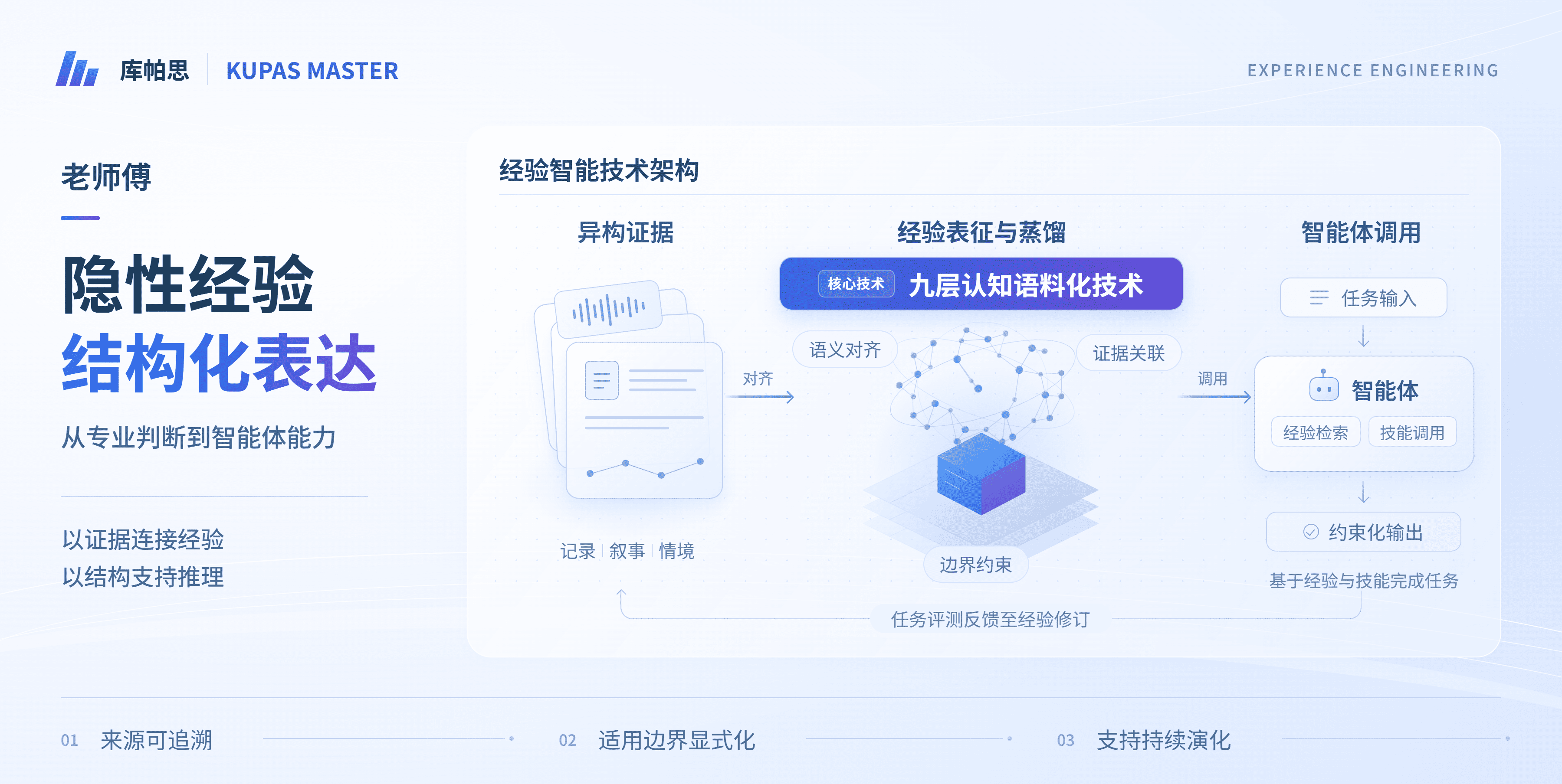}
\caption{System overview of \product and its architecture for using professional experience.}
\label{fig:website-source}
\end{figure}

The \product workflow starts with a clearly scoped task. The task description identifies the practitioner's role, the object of work, the environment, available tools, triggers, and the extent of a single case record. It also defines completion criteria and the decisions the agent is allowed to make. A broad label such as ``industrial maintenance'' is too vague to guide experience collection. A more useful definition names a component and its operating conditions, then limits the task to gathering information, analyzing the problem, and recommending repairs within the authorized scope.

Once the scope is clear, the practitioner and collection assistant prepare a task description and evidence collection plan. They record key observations, judgments, decisions, exceptions, and outcomes. Missing information prompts targeted follow-up. If a diagnosis lacks an important measurement, for example, the next step is to locate the measurement record or clarify it in an interview. This preserves a clear link between facts, judgments, and outcomes.

One practitioner in our sample mediates workplace injury disputes. Their tasks include assessing the nature and causes of an injury, calculating compensation items, negotiating disputes, and reviewing agreements. Relevant evidence includes responsibility findings, medical records, attendance records, and proof of third-party payments. The task specification also defines the mediator's authority: mediation and advice do not replace judicial decisions. Matters outside that authority, or disputes that remain unresolved, follow the appropriate referral procedure. These requirements give experience collection a concrete focus on evidence, reasoning, steps, and conditions of use.

\begin{figure}[htbp]
\includegraphics[width=\linewidth]{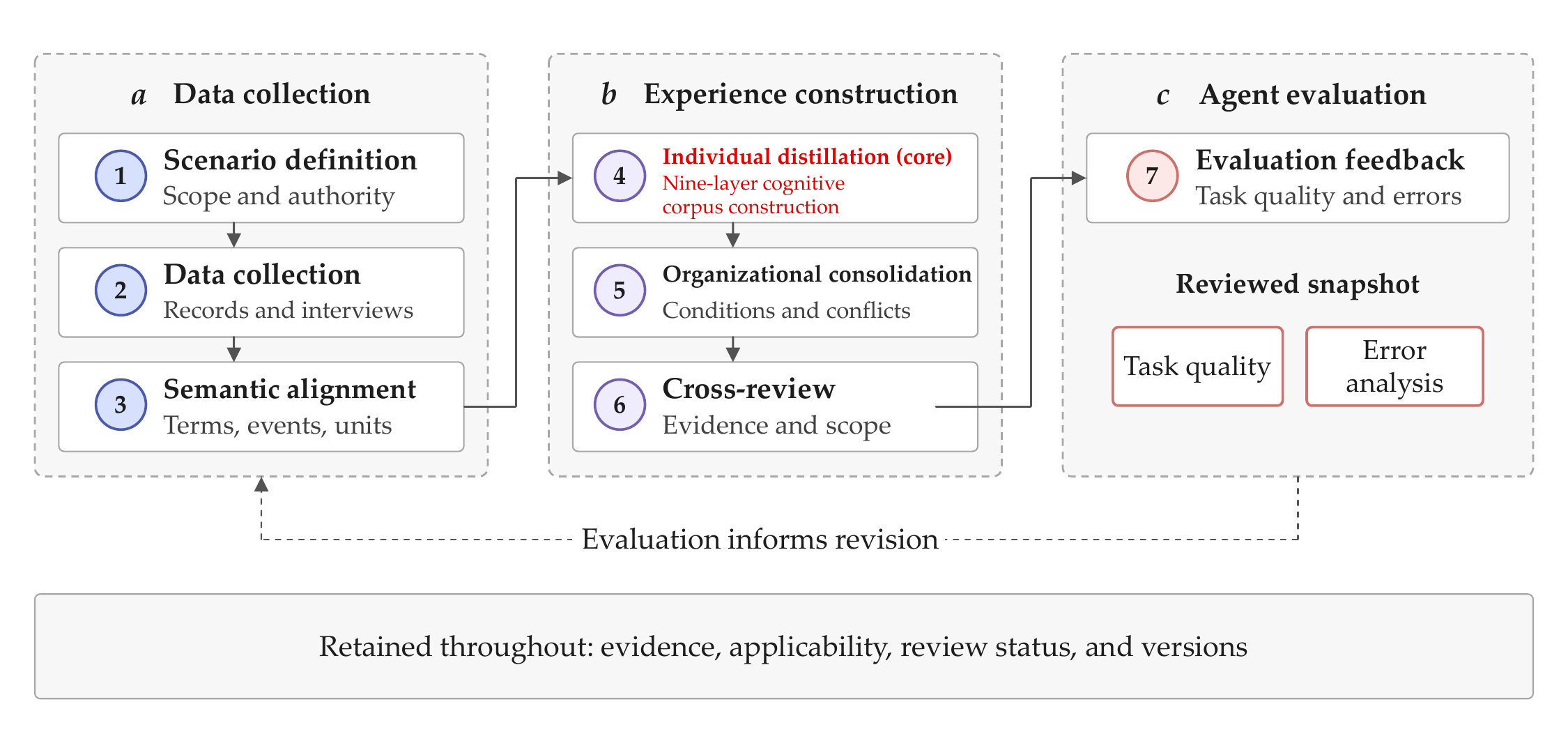}
\caption{The seven-step \product workflow, grouped into data collection, experience construction, and agent evaluation. Solid arrows show processing order. The dashed loop returns evaluation findings to corpus revision. Each stage retains source references, conditions of use, review status, and version identifiers.}
\label{fig:overview}
\end{figure}

Figure~\ref{fig:website-source} shows the platform architecture. Its seven-step workflow is grouped in Figure~\ref{fig:overview}. The platform defines the task, collects evidence, and aligns terms, entities, and events. It then extracts candidate assets from individual cases, compares and consolidates experience across practitioners, and reviews evidence, applicability, and wording. We call this stage \emph{cross-review}. The platform label ``cross-validation'' refers to this content review, rather than $k$-fold cross-validation. Reviewed assets then support agent evaluation. Evaluation findings feed back into collection and revision.

Two distinct records run through the workflow. A \emph{case record} describes what happened during a particular task, including its context, observations, actions, and outcome. An \emph{asset record} captures reusable experience drawn from one or more cases. Their relationship is many-to-many: one case may support several assets, and one asset may draw on several cases. Keeping these records separate preserves the original events while allowing interpretations and generalizations to be revised.

\subsection{Platform responsibilities}
The platform acquires, structures, reviews, selects, and maintains experience. The language model handles task understanding and reasoning, while domain tools obtain external observations or perform authorized actions. Model services connect through a common interface to reduce the effect of model differences on the surrounding workflow.

To make a run traceable and reviewable, its record should include the task specification, model and model version, asset versions, retrieval configuration, and tool permissions. A model change should trigger evaluation on the same fixed tasks with other conditions held constant. Section~\ref{sec:deployment} describes deployment and implementation configurations. The next sections explain how nine-layer cognitive corpus construction supports representation, corpus building, consolidation, and agent integration.

\section{Structured Representation of Expert Experience}
\label{sec:representation}
\product represents experience through case records, extraction dimensions, and asset types. These describe the task process, capture the basis of expert judgment, and organize reusable content, respectively (Figure~\ref{fig:representation}). Their connections are many-to-many: a case can involve several dimensions, and a skill can draw on several asset types.

\begin{table}[!t]
\caption{Case elements and recording requirements. Facts, judgments, and execution states remain distinct, and missing fields are explicit.}
\label{tab:episode}
\begin{tabularx}{\linewidth}{@{}>{\raggedright\arraybackslash}p{0.13\linewidth}XX@{}}
\toprule \lsftablehead{Element} & \lsftablehead{Content} & \lsftablehead{Distinctions to retain}\\\midrule
Context & Task, role, object, environment, time, and available resources. & Conditions observed in this case versus conditions assumed for reuse.\\
Cues & Signals, measurements, statements, and changes noticed during work. & Direct observations versus reported or inferred observations.\\
Judgment & Interpretations, supporting reasons, alternatives, and uncertainty. & Practitioner accounts versus model-generated explanations.\\
Action & Selected operations, order, parameters, and stopping conditions. & Planned, attempted, and completed actions.\\
Boundaries & Prohibitions, scope, missing prerequisites, and referral conditions. & Mandatory constraints versus preferences or usual practice.\\
Outcome & Immediate results, later verification, and unresolved effects. & Expected, reported, and independently verified outcomes.\\
\bottomrule
\end{tabularx}
\end{table}

\reportfigure{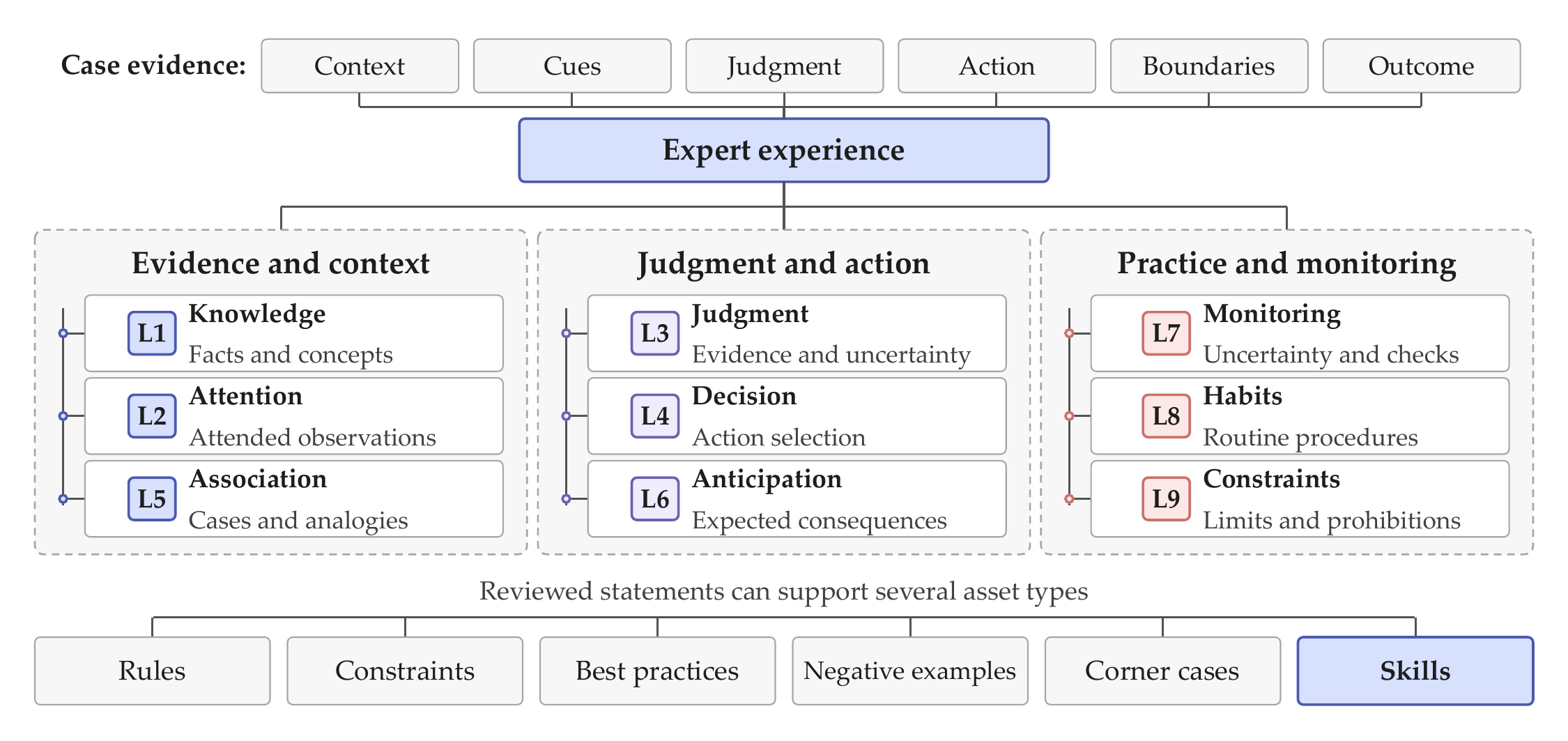}{The nine extraction dimensions, grouped by function. Case elements record the task process, and reviewed statements can become six types of experience asset. Dimensions and asset types have a many-to-many relationship: a skill may combine attention, decision, habit, and constraint information.}{fig:representation}

\subsection{Task cases}
Following the case elements in Table~\ref{tab:episode}, a task case is represented as
\begin{equation}
e=(c,u,j,a,b,o),
\end{equation}
where $c$ is context, $u$ cues, $j$ judgment, $a$ action, $b$ boundaries, and $o$ outcome. These elements provide a common record while allowing different execution orders. Repeated judgments and actions can be recorded as timestamped or partially ordered events, retaining their order and dependencies. Unresolved cases keep their outcome unverified; missing reasons remain explicit and can be added later.

Several independent accounts of the same event can coexist. If practitioners disagree about an observation or fact, each account retains its source rather than being merged into a single certain statement. Earlier failed actions also remain in the record even when a later action solves the problem. They show how feedback changed the practitioner's judgment or strategy. Keeping only the successful path would lose that part of the experience.

For workplace injury mediation, the context might be an injured construction worker requesting compensation mediation. Cues include responsibility findings, medical records, proof of employment, and insurance documents. Judgment concerns the relationship between work injury benefits and third-party payments. Actions include compensation calculations, explanations of policy and responsibility, negotiation, and agreement review. Boundaries specify that the mediator cannot replace a judicial decision and must refer unresolved disputes to arbitration or litigation. An outcome may be an agreement or a record of unresolved issues and completed referral. The six fields let reviewers examine facts, professional judgments, actions, authority, and results separately.

\subsection{The nine-layer framework and extraction dimensions}
The nine-layer cognitive framework supplies a common vocabulary for extraction and annotation (Table~\ref{tab:dimensions}). L1--L9 follow the platform's established terminology and denote dimensions that work together and can overlap. Figure~\ref{fig:representation} groups them into evidence and context, judgment and action, and practice and monitoring. A single statement can span several dimensions or groups.

The dimensions guide questions such as what a practitioner noticed, which evidence supported a judgment, what action followed, and how feedback changed the next step. Together with the asset types, they form the complete experience configuration evaluated on professional tasks.

\begin{table}[!t]
\caption{Extraction questions and representative objects in the nine-layer framework. Outputs become assets after evidence and applicability review.}
\label{tab:dimensions}
\begin{tabularx}{\linewidth}{@{}p{0.055\linewidth}>{\raggedright\arraybackslash}p{0.16\linewidth}XX@{}}
\toprule \lsftablehead{ID} & \lsftablehead{Dimension} & \lsftablehead{Extraction question} & \lsftablehead{Representative objects}\\\midrule
L1 & Knowledge & Which facts and concepts were used? & Entities, terms, relations, and domain materials.\\
L2 & Attention & Which observations received priority? & Diagnostic cues, ignored distractions, and shifts of focus.\\
L3 & Judgment & What supports this interpretation? & Conditional judgments, thresholds, alternatives, and uncertainty.\\
L4 & Decision & How was an action selected? & Prerequisites, selection criteria, and stopping conditions.\\
L5 & Association & Which other cases or concepts were relevant? & Analogies, recalled precedents, and links across cases.\\
L6 & Anticipation & What consequences were expected? & Predicted effects, time horizons, and follow-up checks.\\
L7 & Monitoring & When was another check needed? & Knowledge gaps, uncertainty, and reasons to revisit a judgment.\\
L8 & Habits & Which procedures recur across cases? & Repeated action sequences, communication routines, and preferences.\\
L9 & Constraints & What must not be done? & Prohibitions, scope limits, and referral conditions.\\
\bottomrule
\end{tabularx}
\end{table}

Attention extraction requires direct or indirect evidence of what the practitioner actually attended to, such as an explanation, inspection sequence, or activity log. A model's attention weights alone do not establish a practitioner attention pattern~\citep{jain2019attention}. Anticipation records pair a prediction with its horizon and later outcome, separating prior judgment from subsequent fact. Repeated behavior supports a habit description but does not make it mandatory. Monitoring records uncertainty about judgments, evidence sufficiency, and applicability, preserving the limits on use.

\subsection{Nine-layer cognitive corpus construction}
\label{sec:cognitive-corpusization}
\begin{figure}[!t]
\centering
\includegraphics[width=\linewidth]{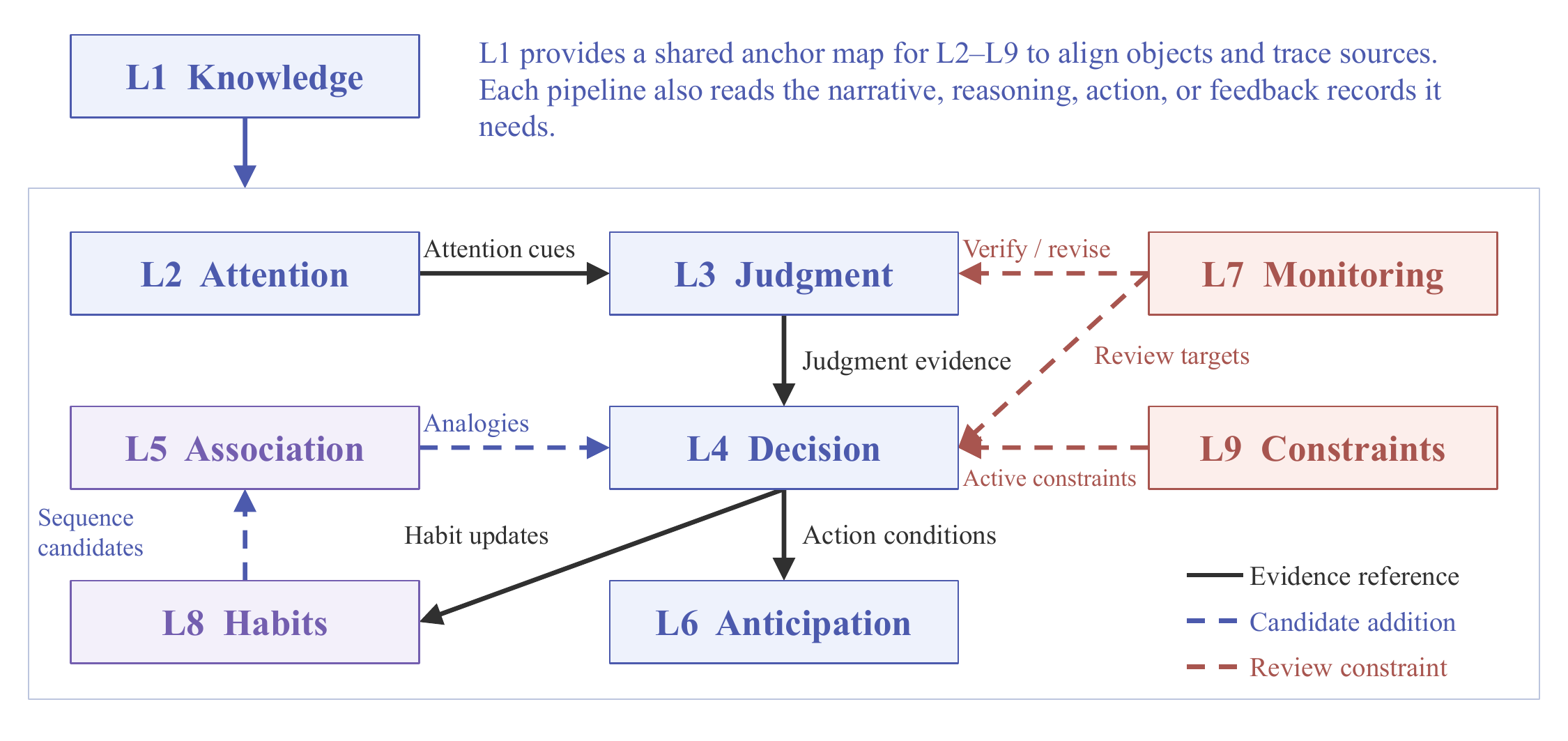}
\caption{Shared dependencies and cross-layer links among the nine extraction pipelines. The arrow from L1 to the L2--L9 group denotes a shared anchor map for object identity and source tracing. Each pipeline also reads its own required source records. Other arrows denote evidence references, candidate additions, or review constraints supported by sources.}
\label{fig:cognitive-pipeline-mapping}
\end{figure}

This section explains how the nine dimensions turn factual accounts, reasoning, and action records into traceable experience units. The approach organizes linked extraction pipelines over narrative materials, explicit reasoning, action logs, and feedback, using entity-relation and event-dependency graphs. It first aligns events across sources in time, merges duplicate records only when their objects and meanings are compatible, and keeps conflicting accounts separate. Changes in meaning, task phase, or feedback state define cognitive segments with common fields for objects, evidence, actions, and outcomes. A segment may enter several pipelines. Figure~\ref{fig:cognitive-pipeline-mapping} shows their shared dependencies and cross-layer links. The methods below are selected according to the available data. Scores and thresholds require calibration on annotated data, and all outputs begin as candidates for review.

\paragraph{L1: Knowledge.}
Coreference resolution, entity linking, and relation classification align concepts, objects, and relations with a graph. For segment $c_i$ and graph version $\nu$, let the anchor mapping be $\mathcal M_\nu(c_i)=(E_i,R_i,\Lambda_i)$, where $E_i$ and $R_i$ contain entities and relations, and $\Lambda_i$ records source locations and event times. Changing entity attributes retain temporal versions, and uncertain links remain candidates. Explicit references, use in judgments or actions, and retractions distinguish a mere mention from evidence of use. Frequency ranking does not automatically remove rare but critical knowledge. The outputs are knowledge units, attribute versions, and anchor mappings. L2--L9 share these mappings for object alignment and source tracing while also reading their own required records.

\FloatBarrier

\paragraph{L2: Attention.}
Attention cues in language, dependency parsing, and inspection or selection logs identify focus objects, yielding a time-ordered record $\mathcal F=\{(t_k,f_k,\ell_k)\}_{k=1}^{m}$. Here, $f_k$ is a focus object or set of objects and $\ell_k$ a source anchor. A transition is recorded only when both adjacent observations are valid. Descriptions and selection changes can then calibrate a switching score; unobserved features remain missing. Across cases, attention frequency is measured relative to opportunities to observe the object in comparable tasks, and patterns are summarized after review. Outputs retain focus objects, transition paths, and evidence types. An unmentioned object is not automatically an ignored object, and model attention weights do not replace practitioner evidence.

\paragraph{L3: Judgment.}
Explicit reasoning provides judgment targets, supporting and opposing evidence, and exclusions. Anchors connect conditions to conclusions, and path compression retains intermediate premises. For a record $x$ whose prerequisites hold, exclusions do not apply, and evidence can be scored, let $q_r(x)\in[0,1]$ be the calibrated support score for rule $r$. A review recommendation can be written as
\begin{equation}
\operatorname{status}_r(x)=
\begin{cases}
\text{supported}, & q_r(x)\geq\tau_{+},\\
\text{review needed}, & \tau_{-}\leq q_r(x)<\tau_{+},\\
\text{do not accept this judgment}, & q_r(x)<\tau_{-},
\end{cases}
\qquad 0\leq\tau_{-}<\tau_{+}\leq1.
\label{eq:cog-judgment-status}
\end{equation}
Support is not a probability of correctness. Missing evidence does not receive a zero score, and an opposite conclusion requires its own evidence. With explicit likelihoods and priors, Markov chain Monte Carlo (MCMC) can sample posterior distributions over numerical domain thresholds. Review thresholds $\tau_{-}$ and $\tau_{+}$ instead require calibration on labeled examples. Low support alone does not revoke an asset. Change-point detection applies only to ordered data that meet its piecewise distribution assumptions. The output is a judgment rule with applicability, validity and failure boundaries, and evidence references.

\paragraph{L4: Decision.}
The pipeline extracts candidate actions, reasons for rejection, and trigger, stopping, and fallback conditions, then links actions to operation nodes. Let $\mathcal C(x)$ be the candidates in context $x$. Let $K(x,a)$ mean that relevant facts and local-rule applicability have been checked, $F(x,a)$ denote local feasibility, and $H_r(x,a)$ denote satisfaction of an applicable local restriction $r\in\mathcal R_{\mathrm{loc}}(x)$. Then
\begin{equation}
\mathcal C_{\mathrm{loc}}(x)=
\left\{a\in\mathcal C(x)\;\middle|\;
K(x,a)\land F(x,a)\land\bigwedge_{r\in\mathcal R_{\mathrm{loc}}(x)}H_r(x,a)\right\}.
\label{eq:cog-action-filter}
\end{equation}
Only actions with all conditions confirmed enter this set. Unknown cases await review. L9 jointly checks required actions, mutual exclusions, and timing across the full plan, so local acceptance does not establish overall compliance. When $|\mathcal C(x)|>0$, the ratio $\rho(x)=1-|\mathcal C_{\mathrm{loc}}(x)|/|\mathcal C(x)|$ measures the reduction to the confirmed set. Excluded candidates are not necessarily infeasible. A single remaining candidate still needs evidence for selection; an empty set calls for clarification or referral.

Condition-action pairs across cases can train a pruned C4.5 decision tree for testing on held-out cases. With explicit states, actions, transitions, and reward features, maximum entropy inverse reinforcement learning can estimate a candidate reward function. Textual comparisons first establish a partial order over criteria. Experts can complete pairwise comparisons and check consistency before applying the Analytic Hierarchy Process (AHP) to compute weights. Outputs distinguish candidate, planned, selected, and performed actions. A candidate reward function is not treated as the uniquely true preference.

\paragraph{L5: Association.}
Explicit association statements identify a source concept $u$, target concept $v$, and a connecting cue. Graph shortest-path distance $d_G(u,v)$ and cosine similarity between nonzero embeddings, $\cos(\mathbf h_u,\mathbf h_v)$, describe structural distance and semantic similarity. Language cues, the basis of an analogy, and shared attributes help filter candidates while preserving graph versions and scoring evidence. PrefixSpan can mine frequent subsequences from operation logs, but a pattern is labeled as a practitioner association only with verbal or other process evidence. Outputs include supported associations and links awaiting verification, which can suggest explanations or alternative actions. These associations still need careful interpretation: a missing graph edge does not establish a cognitive leap, and association does not establish causation.

\paragraph{L6: Anticipation.}
The pipeline extracts the current state, candidate action, expected consequence, and time horizon. It creates prediction-event nodes and freezes the information, rule version, and timestamp available at prediction time. To verify a conditional prediction, it first checks that the action and trigger actually occurred, then matches the object, outcome definition, and observation window. An unexecuted plan is not paired directly with an observed outcome. For a verified pair $i$, let $(\widehat y_i,y_i)$, $(\widehat\kappa_i,\kappa_i)$, and $(\widehat t_i,t_i)$ denote predicted and actual numerical values, categories, and event times. Record errors by type:
\begin{equation}
e_i^{\mathrm{num}}=\widehat y_i-y_i,\qquad
e_i^{\mathrm{cat}}=\mathbf 1[\widehat\kappa_i\ne\kappa_i],\qquad
e_i^{\mathrm{time}}=\widehat t_i-t_i.
\label{eq:cog-forecast-errors}
\end{equation}
The indicator $\mathbf1[\cdot]$ is 1 when its condition holds and 0 otherwise. Errors retain their own units and are not added across types. If an event does not occur within a complete observation window, an occurrence prediction receives a negative outcome and timing error is undefined. An incomplete window or record remains unverified. Probabilistic predictions require calibration checks on independent samples with sufficient follow-up labels. Physical tasks may also retain states and contact or support relations. Counterfactual analysis requires particular care: causal mechanisms and identification assumptions must be explicit, and differences in simulated outcomes are not verified action effects. Outputs contain the prediction, its valid horizon, and verification records.

\paragraph{L7: Monitoring.}
Sequence labeling identifies uncertainty, requests for verification, revisions, and statements about role limits, linking each to a proposition $p$ and its evidence. The monitoring state $z_t(p)$ records transitions such as doubt, verification requests, retraction, and reinstatement after checking, together with the evidence that triggered them. A topic change alone is not a state change. Stated confidence, evidence sufficiency, graph gaps, and role boundaries remain separate. Linguistic confidence is not converted directly into a probability of correctness. Outputs identify knowledge gaps, unresolved checks, and review conditions for the judgment and decision pipelines.

\paragraph{L8: Habits.}
Logs are grouped by practitioner, task type, and comparable context, then normalized into object-action sequences. For $N>0$ valid case sequences $S_i$, define the case-level support of a nonempty pattern $s$ and a normalized Levenshtein distance with unit insertion, deletion, and substitution costs:
\begin{equation}
\operatorname{supp}(s)=\frac1N\sum_{i=1}^{N}\mathbf1[s\preceq S_i],\qquad
d_{\mathrm{norm}}(S_i,S_j)=\frac{d_{\mathrm{edit}}(S_i,S_j)}{\max\{1,|S_i|,|S_j|\}}.
\label{eq:cog-habit-statistics}
\end{equation}
Here $s\preceq S_i$ denotes an ordered, not necessarily contiguous subsequence, and $|S_i|$ is sequence length. Repeated occurrences within one case count once. Stability checks consider recurrence over time, sample size, and exceptions alongside support and distance. Direct quotations can become wording templates after object and role slots are introduced and semantically equivalent expressions are grouped. Outputs include atomic operations, compound procedures, and optional wording. Performed actions verified in L4 logs can support patterns across cases, and sequence candidates can inform L5, where connecting cues or process evidence are still required. These links reuse corpus content; they are not an online execution loop. An observed habit does not by itself establish competence or a mandatory requirement.

\paragraph{L9: Constraints.}
Negation-scope parsing and checks of normative sources recover prerequisites, prohibitions, required actions, responsibility boundaries, and alternatives. Within a task and agreed time window, an action identifier $a$ includes its actor, object, and occurrence. Boolean variable $X_a$ records whether the action occurs, with time $T_a$ added when needed. Let $P_r(x)$ be the applicability condition of rule $r$, and $\mathcal R^{-}$ and $\mathcal R^{+}$ the prohibition and obligation sets. One constraint formula is
\begin{equation}
\Phi_x=\Phi_{\mathrm{task}}(x)\land
\bigwedge_{r\in\mathcal R^{-}}\!\bigl(P_r(x)\Rightarrow\neg X_{a_r}\bigr)
\land\bigwedge_{r\in\mathcal R^{+}}\!\bigl(P_r(x)\Rightarrow X_{a_r}\bigr).
\label{eq:cog-norm-constraints}
\end{equation}
Here $a_r$ is the action governed by rule $r$, and $\Phi_{\mathrm{task}}$ encodes verified facts, prerequisites, mutual exclusions, and timing. Time constraints apply only to performed actions, and different scopes are modeled separately. A satisfiability modulo theories (SMT) solver can check $\Phi_x$. Satisfiability means only that the encoded conditions admit a consistent solution. Unknown applicability still requires evidence and review, even if the formula is satisfiable; a solver assignment is not an observed fact. Unsatisfiable cases and cases for which the solver returns \texttt{unknown} go to human review. Active hard constraints must be satisfied before risk signals determine whether to request review, reduce output detail, or suppress output. Scores cannot override hard constraints. Missing logs do not prove compliance. Outputs retain their basis, scope, and review status.

All nine pipelines save case and segment identifiers, graph anchors, content, conditions, sources, review status, and versions. Cross-layer mapping first checks task, object, and time scope, then uses explicit references, temporal proximity, and content agreement to find and rank candidate links. Source evidence determines whether a link is an evidence reference, candidate addition, or review constraint. Exceeding a score threshold does not by itself establish a relation type or a causal claim. Reviewed reusable content is organized into six asset types while retaining its case and graph links. Quality control checks structure, evidence, conditions, and task use. It does not require every segment to cover all nine dimensions or treat more links as higher quality. The next section discusses the framework's functional analogies with brain systems.

\subsection{Functional correspondences with brain systems}
\label{sec:brain-mapping}
The nine-layer approach provides a common vocabulary for extracting, reviewing, and reusing experience from heterogeneous sources. Its design draws inspiration from the organization of cognitive functions in neuroscience. These correspondences guide extraction questions and give the dimensions a coherent functional basis.

\begin{figure}[htbp]
\includegraphics[width=\linewidth]{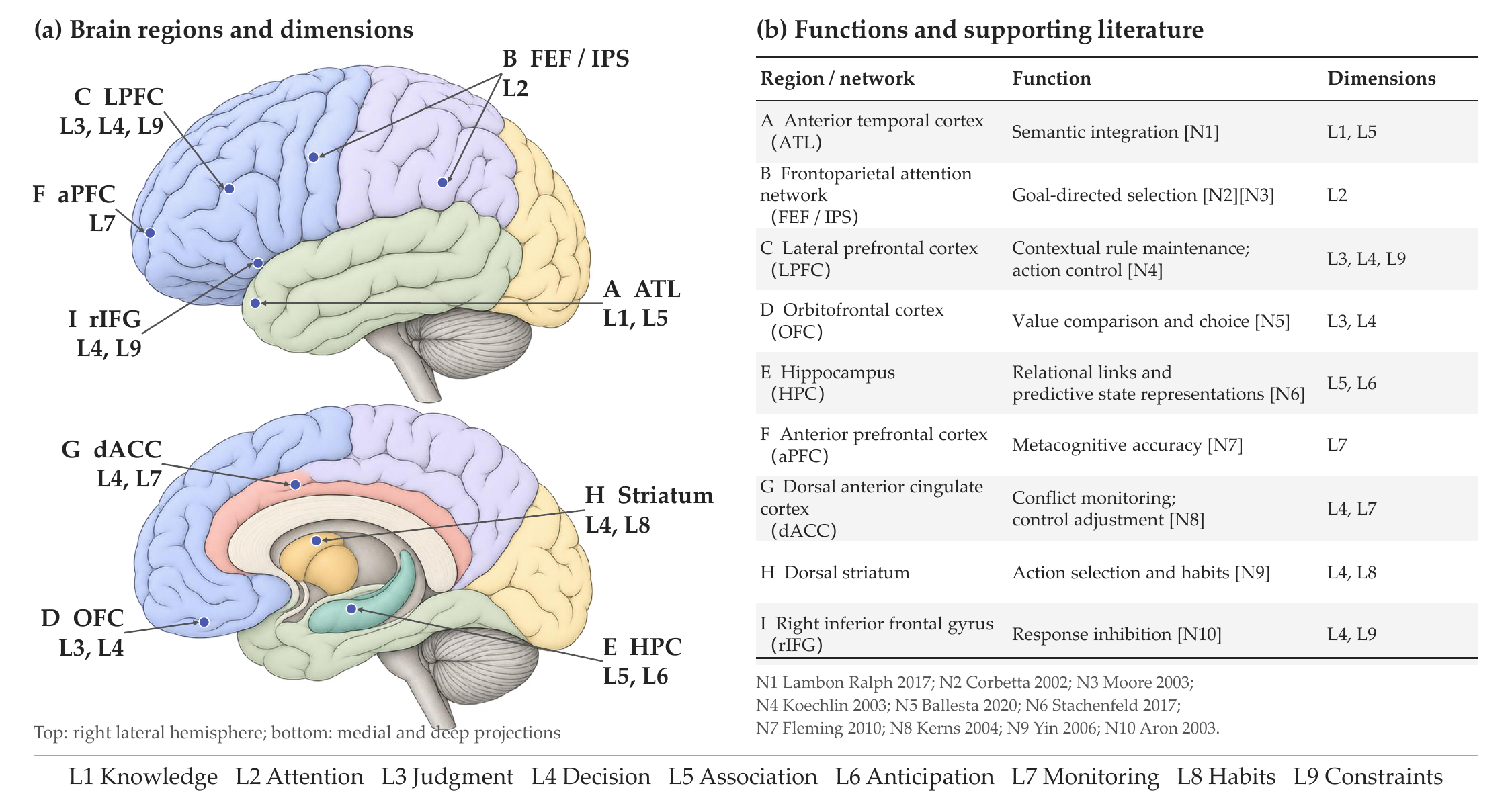}
\caption{Many-to-many functional correspondences between brain systems and the nine-layer framework, based on neuroscience studies. The upper drawing shows the right lateral hemisphere; the lower drawing projects medial and deep structures. Letters match the table rows, and arrows indicate approximate locations. References: N1~\citep{lambonralph2017semantic}; N2~\citep{corbetta2002attention}; N3~\citep{moore2003gating}; N4~\citep{koechlin2003control}; N5~\citep{ballesta2020values}; N6~\citep{stachenfeld2017predictive}; N7~\citep{fleming2010introspection}; N8~\citep{kerns2004conflict}; N9~\citep{yin2006habits}; N10~\citep{aron2003stopsignal}.}
\label{fig:brain-mapping}
\end{figure}

Figure~\ref{fig:brain-mapping} connects L1--L9 to functions described in the neuroscience literature. This is an analytical framework for organizing expert experience, with many-to-many functional references to cooperating brain regions and networks. The anterior temporal lobe and distributed association cortex support semantic representation and integration, providing references for L1 knowledge and L5 association~\citep{lambonralph2017semantic}. Predictive representations in hippocampal systems connect existing relationships to possible successor states, informing L5 association and L6 anticipation~\citep{stachenfeld2017predictive}. The frontal eye fields and intraparietal sulcus participate in goal-directed selection, which resembles the selective processing considered in L2 attention~\citep{corbetta2002attention,moore2003gating}.

The lateral prefrontal cortex contributes to context-dependent cognitive control, while the orbitofrontal cortex represents and compares values. These functions provide references for judgment, decision, and rule maintenance~\citep{koechlin2003control,ballesta2020values}. Anterior prefrontal and anterior cingulate functions related to self-monitoring, conflict detection, and adjustment offer analogies for monitoring experience~\citep{fleming2010introspection,kerns2004conflict}. Striatal circuits involved in goal-directed behavior and habitual responses inform L4 and L8~\citep{yin2006habits}. Right inferior frontal involvement in response inhibition provides a partial reference for stopping and control in L4 and L9~\citep{aron2003stopsignal}. Together, these functional analogies organize experience extraction, while explicit records, rules, and review govern permissions and applicability.

\subsection{Six experience asset libraries}
Experience assets package reusable content from case records, practitioner statements, and authoritative procedures for a defined task. The six core types are rules, constraints, best practices, negative examples, corner cases, and skills. ``Negative examples'' follows the platform's terminology for failures, ineffective practices, and corrections; it does not mean negative-class training examples. The corner-case library records unusual situations in which routine rules may fail or need adjustment, including cases far from a numerical threshold.

Dimensions and asset types are connected many-to-many. L1 facts, entities, and relations generally remain in case records or the shared graph. L6 predictions stay with their horizons and observed outcomes and may also become applicability conditions for a rule, skill, or case. Several dimensions can therefore support one asset.

Some content can be included as extensions. Wording styles and role preferences may be optional skill settings. Knowledge gaps remain explicit metadata that trigger collection or review, without requiring another top-level library. Common fields define the core assets, while extensions support domain-specific types in commercial deployments.

\begin{table}[!t]
\caption{Asset types, reusable content, and common source dimensions. Assets can combine information across dimensions.}
\label{tab:assets}
\begin{tabularx}{\linewidth}{@{}>{\raggedright\arraybackslash}p{0.16\linewidth}X>{\raggedright\arraybackslash}p{0.18\linewidth}@{}}
\toprule \lsftablehead{Asset} & \lsftablehead{Reusable content} & \lsftablehead{Common dimensions}\\\midrule
Rules & Scoped conditions and recommended judgments or actions, including exceptions. & L3, L4\\
Constraints & Prohibitions or required prerequisites, their basis, and allowed alternatives. & L7, L9\\
Best practices & Supported successful procedures and conditions for considering reuse. & L1, L4, L5, L6\\
Negative examples & Inappropriate actions or adverse outcomes, context, and reviewed corrections. & L3, L4, L7, L9\\
Corner cases & Unusual contexts where a routine rule may fail or need adjustment. & L2, L3, L5, L7\\
Skills & Callable procedures with inputs, outputs, dependencies, checks, and evidence. & L2, L3, L4, L8, L9\\
\bottomrule
\end{tabularx}
\end{table}

The sample used in this report contains 13,113 organizational assets: 3,550 rules, 2,892 best practices, 1,955 skills, 1,948 corner cases, 1,623 constraints, and 1,145 negative examples. Rules and best practices guide judgments and operations. Constraints, negative examples, and corner cases supply applicability conditions, risk information, and exceptions. Skills organize this experience into callable task procedures.

Among the 53 questions with individual tool-call records, C retrieved rules in 46 runs, corner cases in 35, best practices in 33, constraints in 28, and negative examples in 18. It called skills in 46 runs. These records show how libraries are combined during retrieval, reasoning, and execution.

Negative examples and corner cases serve different purposes. Negative examples preserve observed mistakes, failures, and corrections to help prevent repetition. Corner cases identify situations that require adjustment even if no failure has yet occurred. Best practices retain both outcomes and conditions for reuse. A skill combines these asset types under a common specification for inputs, steps, state transitions, and outputs.

\subsection{Evidence, review, and uncertainty}
Every transformation should retain the source type: direct observation, practitioner account, or model proposal. A model may suggest missing relations or reconstruct possible reasons from behavior, but these remain candidate explanations for review and are stored separately from observations and practitioner statements. In this report, \emph{distillation} means extracting, organizing, and consolidating experience from materials, rather than training a student model from a teacher.

Review status and confidence are separate. Extraction confidence concerns whether parsing recovered the original statement accurately. Evidence sufficiency concerns whether sources support it. Applicability concerns whether it can be used in a particular task or context. The system stores these separately so reviewers can inspect each question.

An asset may move through candidate, reviewed, active, suspended, and retired states. Review examines content, evidence, and declared scope. Activation also requires deployment and runtime checks. A correction creates a new version with a replacement link to the old one. Even when a generalization is revoked or found invalid, its original cases remain available to trace how the experience changed.

The organizational assets in this sample retain library versions and entry identifiers, allowing references and version relationships to be checked. Evaluation records should also include the corpus, asset versions, and runtime configuration actually used, supporting review and regression tests after changes.

\subsection{Graph structures and requirements for agent use}
Entity graphs organize task objects and relations; event and dependency graphs describe task progression and links between steps. Edge types distinguish temporal order, procedural dependency, association, and explicitly proposed causal hypotheses. Event order within a case can form a directed acyclic structure. Repeated operations or decisions can instead use repeated event instances or a separate state machine. Explicit edge types keep their meanings distinct.

An asset is ready for agent use when a consumer can understand its content, applicability, evidence, and permitted uses, and respond appropriately to missing prerequisites. The interface therefore needs machine-readable fields for task scope, exclusions, sources, review status, and versions. Skills also require inputs, outputs, and execution conditions. Interface validation checks that required fields and constraints are present; task evaluation checks whether the agent selects and uses the right asset in context.

The following sections explain how corpus construction and consolidation produce traceable assets under these requirements. Section~\ref{sec:case} then shows their retrieval and use in recorded workplace injury mediation tasks.

\section{Experience Acquisition and Corpus Construction}
\label{sec:corpus}
\subsection{Collecting heterogeneous evidence}
Experience collection draws on existing materials and additional records of actual work. Existing sources include case files, work logs, incident reviews, training materials, and demonstration videos. Additional collection may involve interviews, task observation, application logs, sensors, audio, and video. The right method depends on the task. An information-heavy review process may already have detailed digital records, whereas physical work often requires observation and follow-up interviews to explain changes in action. Materials should cover key decisions, outcomes, and exceptions. Interviewees should have relevant experience, be able to explain important cues, and provide cases that can be checked.

Process records should preserve the order of work and distinguish information available at the time from information learned later. Interviews can recover omitted observations, rejected alternatives, and conditions under which the usual procedure fails. These accounts should be marked as retrospective. Measurements or observations that remain unavailable should be listed explicitly for later collection.

The authorized sample contains 1,576 files in 13 formats: 567 DOCX, 351 PDF, 172 JSON, 154 Markdown, 131 DOC, and 95 XLSX files, plus text, webpages, images, presentations, and audio. The materials include case files, procedures, standards, training materials, interviews, question-answer records, tabular data, and de-identified records. Processing therefore needs to handle narrative text, tables, and multimodal attachments while keeping related content linked.

\reportfigure{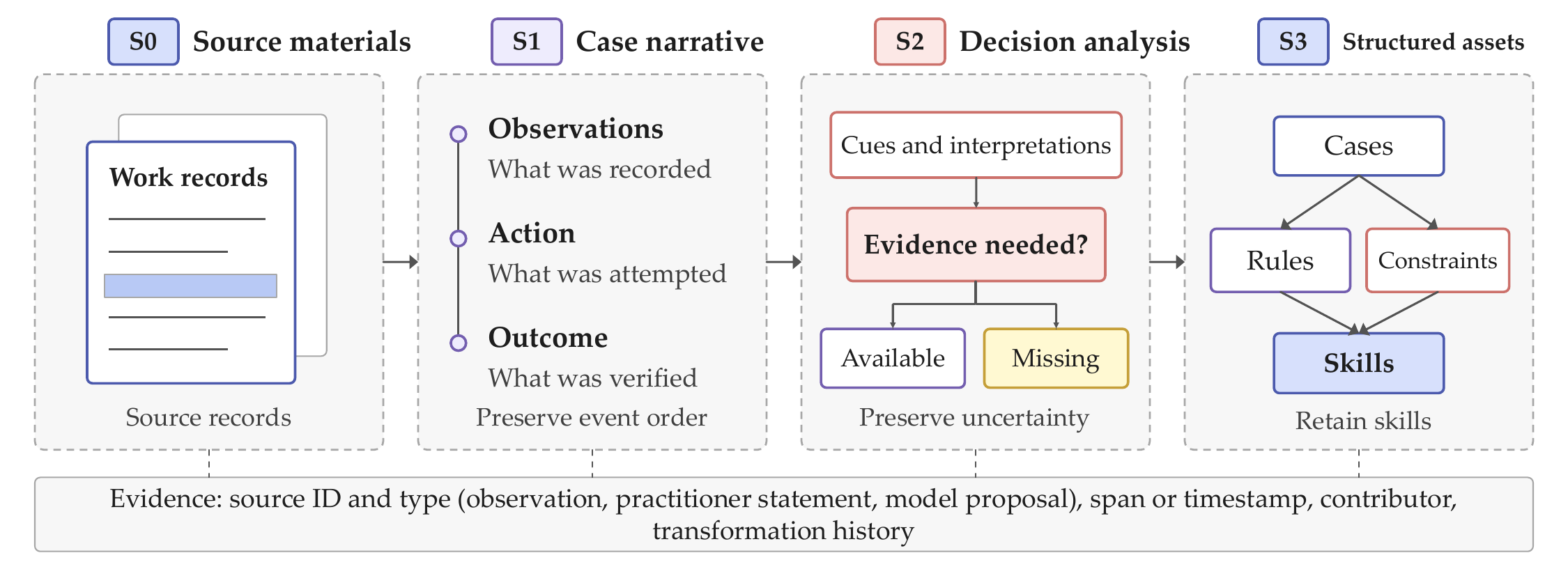}{Four representations in corpus construction. Source materials (S0) become case narratives (S1), which are analyzed for decision elements (S2) and converted into structured candidate records (S3). Dashed links preserve source traceability across stages.}{fig:corpus}

\subsection{Semantic alignment}
The platform retains original wording while normalizing identifiers, terms, event boundaries, and units. Abbreviations, colloquial or local expressions, and ambiguous references require context. A mapping record should preserve the original phrase, normalized term, supporting context, and plausible alternatives that remain unresolved. A phrase such as ``slightly hot'' should become a numerical range only when measurements or a reviewed domain definition support that mapping.

Multimodal observations are aligned with task events and time intervals. Transcriptions and image recognition outputs retain uncertainty and source locations. Sensor records include units, sampling methods, and available calibration information. Because recognition errors can affect later rules and skills, every derived claim should remain traceable to the original record.

The hypertension practitioner provides a concrete example. Their materials comprise two 2024 Chinese hypertension guidelines and XML field definitions for a hypertension follow-up dataset. The guidelines describe diagnosis, monitoring, and follow-up. The XML specifies fields such as systolic and diastolic blood pressure, body mass index, medication adherence, adverse reactions, referral reasons, and the next follow-up date. Semantic alignment connects guideline conditions to these fields, units, and value ranges while retaining guideline versions and source locations. Later follow-up entries can then use consistent fields and refer back to the relevant guideline passages. The XML supplies the data schema; actual follow-up data are entered during subsequent clinical work.

\subsection{Narrative reconstruction and structured extraction}
The S0--S3 stages in Figure~\ref{fig:corpus} separate case reconstruction from experience generalization. S0 retains source materials. S1 organizes the case by roles, observations, actions, and outcomes. S2 identifies decision elements, including cues, explanations, alternatives, plan changes, and uncertainty. S3 converts supported elements into structured records and graphs. All stages remain linked to their sources.

The extractors then apply the nine-layer approach in Section~\ref{sec:cognitive-corpusization}. Knowledge extraction links terms and entities. Attention extraction identifies priorities supported by practitioner statements or behavioral evidence. Judgment extraction recovers conditions and conclusions. Decision extraction records actions, selection criteria, and stopping conditions. Association extraction captures related cases that the practitioner mentions, without treating similarity as causal evidence. Anticipation extraction records predictions and time horizons. Monitoring extraction records when more information, review, or referral is needed. Habit extraction requires repetition across cases. Constraint extraction distinguishes explicit prohibitions from preferences.

An extractor may combine language model prompts, rule parsing, graph processing, and domain rules. If it involves model training, the training data, objective, and results on an independent validation set should also be recorded. Regardless of implementation, candidate assets need structured fields and supporting evidence that a reviewer can inspect.

\begin{table}[!t]
\caption{Core transformation operators and reasons to reject an output or leave it unresolved. Every operator records input and output versions and source locations.}
\label{tab:operators}
\begin{tabularx}{\linewidth}{@{}>{\raggedright\arraybackslash}p{0.24\linewidth}XX@{}}
\toprule
\lsftablehead{Operator and input} & \lsftablehead{Transformation and output} & \lsftablehead{Reject or leave unresolved when}\\\midrule
Term alignment: wording and context & Propose a standard term while retaining aliases, units, and source spans. & Multiple interpretations remain plausible, or a unit conversion lacks support.\\
Claim extraction: case narrative & Separate observations, judgments, and planned actions, each with its own source. & The extracted claim lacks supporting content.\\
Condition preservation: conditional statement & Recover triggers, recommendations, prerequisites, and exceptions as a candidate rule. & Compression changes the action, drops an exception, or alters a prerequisite.\\
Conflict classification: comparable asset pair & Check scope overlap and action compatibility to distinguish agreement, different conditions, and conflict. & Scope overlap is unknown or evidence is insufficient to resolve the conflict.\\
\bottomrule
\end{tabularx}
\end{table}

A model can propose candidates for practitioners to review against these criteria. Each stage retains its input span, proposed output, revisions, and review decision. Consider the statement ``During initial intake, request verification before assigning a cause, except when a separate emergency procedure applies.'' A valid rule must retain both the intake prerequisite and the emergency exception. Dropping either fails the condition-preservation check.

\FloatBarrier
\paragraph{Preserving conditions during transformation.}
For a rule $r$, let $P_r$ denote its prerequisites, $E_r$ its exclusions, and $a_r$ its recommended action. Within a predefined set of task states $\Omega$, its allowed scope is
\begin{equation}
D(r)=\{x\in\Omega:P_r(x)=\mathrm{true}\ \land\ E_r(x)=\mathrm{false}\}.
\label{eq:scope}
\end{equation}
Unknown values do not establish applicability. For a source rule $r_s$ and an extracted rule $r_e$, define the added and lost scope as
\begin{equation}
B^+(r_e,r_s)=D(r_e)\setminus D(r_s),\qquad
B^-(r_e,r_s)=D(r_s)\setminus D(r_e).
\label{eq:scope-change}
\end{equation}
Normalization should leave both sets empty while preserving the action's meaning and required exceptions. A nonempty $B^+$ introduces unsupported situations; a nonempty $B^-$ removes situations covered by the source. An intentional scope change requires a separate review. These sets define the consistency target, which targeted task tests check through conditions and applicability.

\subsection{Quality control and corpus versions}
Quality checks operate at three levels. Structural checks verify identifiers, required fields, valid references, and types. Semantic review checks whether evidence supports the content and its stated scope. Task review checks whether an asset helps make a decision or follow a procedure without hiding missing prerequisites. Review by other practitioners can expose ambiguities and assumptions that the original contributor takes for granted, and help distinguish personal habits from procedures suitable for wider use.

Duplicate detection should consider derivation as well as text similarity. Several summaries of the same event are not independent evidence. Each experiment or deployment should fix a data version and record its source files, processing programs, active assets, and review decisions. Earlier versions must remain available after updates so that reported results can be checked.

\paragraph{Review status of the sample materials.}
The authorized source materials occupy about 1.73 GB. The file inventory marks 1,564 files as Approved and 4 as Rejected, with 8 lacking a review status. These are file counts. Task sets, corpus versions, and processing configurations are recorded separately for evaluation review.

In the comparative evaluation, adding raw-corpus retrieval raised the composite mean from 70.63 in A to 79.75 in B. The evidence sufficiency and accuracy score rose from 62.18 to 77.27. These results demonstrate that parsing, indexing, and retrieving source materials strengthen evidence-grounded answers and improve task performance.

\section{Organizational Experience Consolidation}
\label{sec:consolidation}
\subsection{From individual accounts to organizational assets}
Practitioners may agree, differ because they work under different conditions, or recommend different actions under the same conditions. Simply pooling their libraries can hide these distinctions. Before combining assets, \product compares their applicability, supporting evidence, and review status.

The platform first aligns task identifiers, entities, terms, and asset types. Semantic similarity and scope overlap then identify entries worth comparing. Similarity is a way to find candidates, not sufficient evidence for a merge. The comparison asks whether the assets address the same decision, whether their conditions can hold together, and whether their judgments or actions are compatible.

\reportfigure{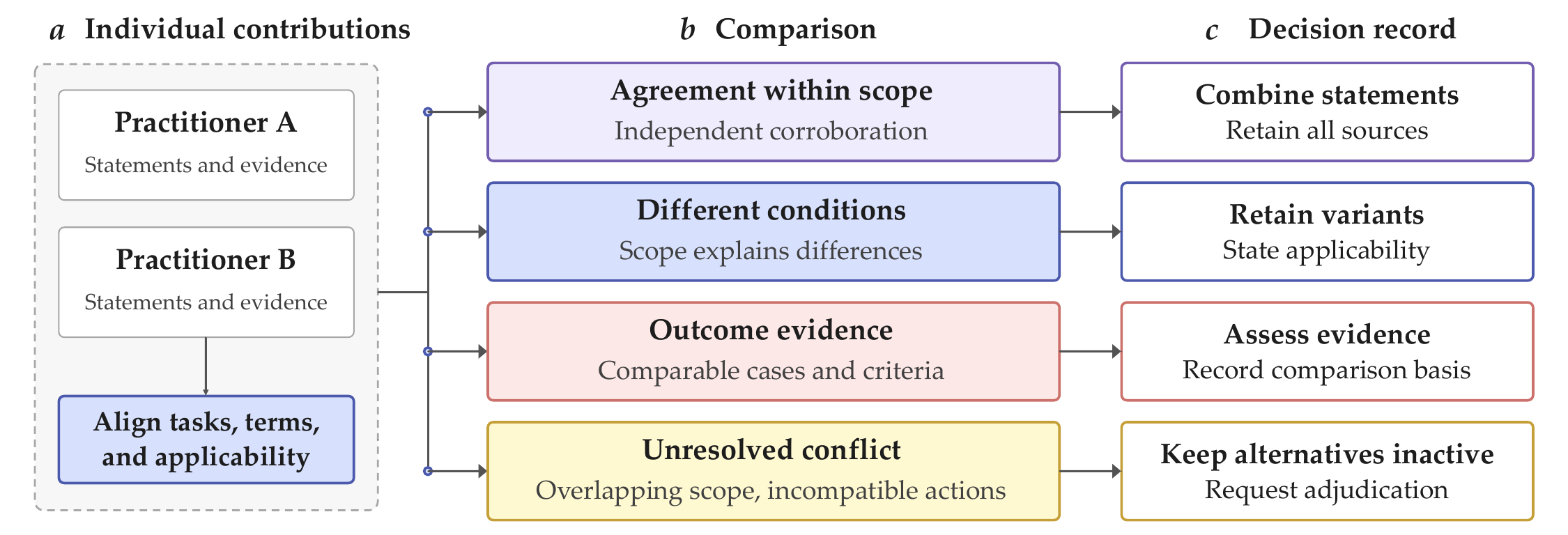}{Four outcomes of organizational consolidation and their evidence requirements. Agreement retains independent support. Context-dependent alternatives remain separate. Outcome comparison requires comparable cases. Unresolved contradictions remain inactive until review. Consolidated assets retain both supporting evidence and conditions of use.}{fig:consolidation}

\subsection{Agreement, different conditions, and conflicts}
Figure~\ref{fig:consolidation} shows four consolidation outcomes. When accounts agree under the same conditions, the combined entry retains all independent sources and their derivation links. When different conditions explain different recommendations, both procedures remain available with explicit conditions. A routine inspection and a procedure for abnormal operation should not be averaged into one sequence.

When comparing alternatives, the platform organizes historical outcomes by case difficulty, operating environment, resources, and outcome definition. Historical records preserve practical evidence; controlled comparisons help assess differences between procedures under comparable conditions.

An unresolved contradiction remains visible as linked alternatives for review or adjudication. It should not become an unconditionally active rule. Reviewers may narrow its scope, request more cases, or decide that neither alternative is ready for use. They also distinguish formal requirements from personal preferences. A requirement grounded in a standard or procedure cannot be settled by a simple vote or a compromise with preferences.

Using the scope in Eq.~\eqref{eq:scope}, define the overlap and conflict regions of two rules as
\begin{equation}
O_{ij}=D(r_i)\cap D(r_j),\qquad
C_{ij}=\{x\in O_{ij}:\operatorname{compatible}(a_i,a_j,x)=\mathrm{false}\}.
\label{eq:conflict}
\end{equation}
Compatibility depends on task order, resources, and procedural requirements. Different actions need not conflict. A nonempty $C_{ij}$ requires adjudication. Unknown overlap or compatibility remains unresolved. If the overlap is empty, both rules can be retained within the checked task scope. Domain review is still needed to establish the relevant conditions and action compatibility.

For example, one intake procedure may apply when there is no emergency and another when there is an emergency. If emergency status is known, their scopes are disjoint and both can be kept. If both apply to non-emergency intake but require incompatible next actions, the platform records a conflict with the asset versions, shared conditions, and unresolved conclusion. If emergency status is unknown, the agent first asks for the missing information.

The workplace injury mediation sample also requires separate procedures for ordinary compensation calculations, third-party commercial insurance, and statutory work injury insurance. The platform must record the insurance type, liable party, eligible compensation items, and mediation authority. A record that merely says ``insurance exists'' leaves applicability unknown. The agent must verify the type and beneficiary relationship before using the relevant experience. Separate conditions make it possible to retrieve the procedure that matches the case.

\subsection{Cross-review and version management}
Other practitioners can review candidate assets and how they were produced. They check sources, conditions, exceptions, consistency with other assets, and possible misuse. Tests that change one important condition, such as whether a measurement is available or referral is required, help check whether applicability is clear.

Consolidation produces both organizational assets and decision records: which inputs were merged, which were kept separate, why, and what remains unresolved. Each update creates a new version. Dependencies identify skills that need revalidation after a rule changes. Version records preserve original contributions and the exact assets used in a run.

\paragraph{Review records in the sample.}
In the sample collected for this report, organizational assets are stored by library and version. Records for 8 practitioners contain 37 completed cross-review jobs across the six libraries, covering 818 candidates. ``Completed'' means that comparison, scoring, and record storage finished. Human confirmation is recorded separately: 9 candidates had completed human verification and 331 were marked as awaiting it. Separating automated results from human review status makes the processing state of each candidate clearer.

For the workplace injury mediator, two rule-library review jobs used the same source version. Each involved 37 source assets and 21 candidates, with processing status retained. These records make experience traceable from its source through review and consolidation. The three-configuration evaluation demonstrates the value of the resulting organizational assets and skills in professional task execution.

\section{Skill Construction and Agent Integration}
\label{sec:skills}
\subsection{Skill invocation and execution requirements}

\reportfigure{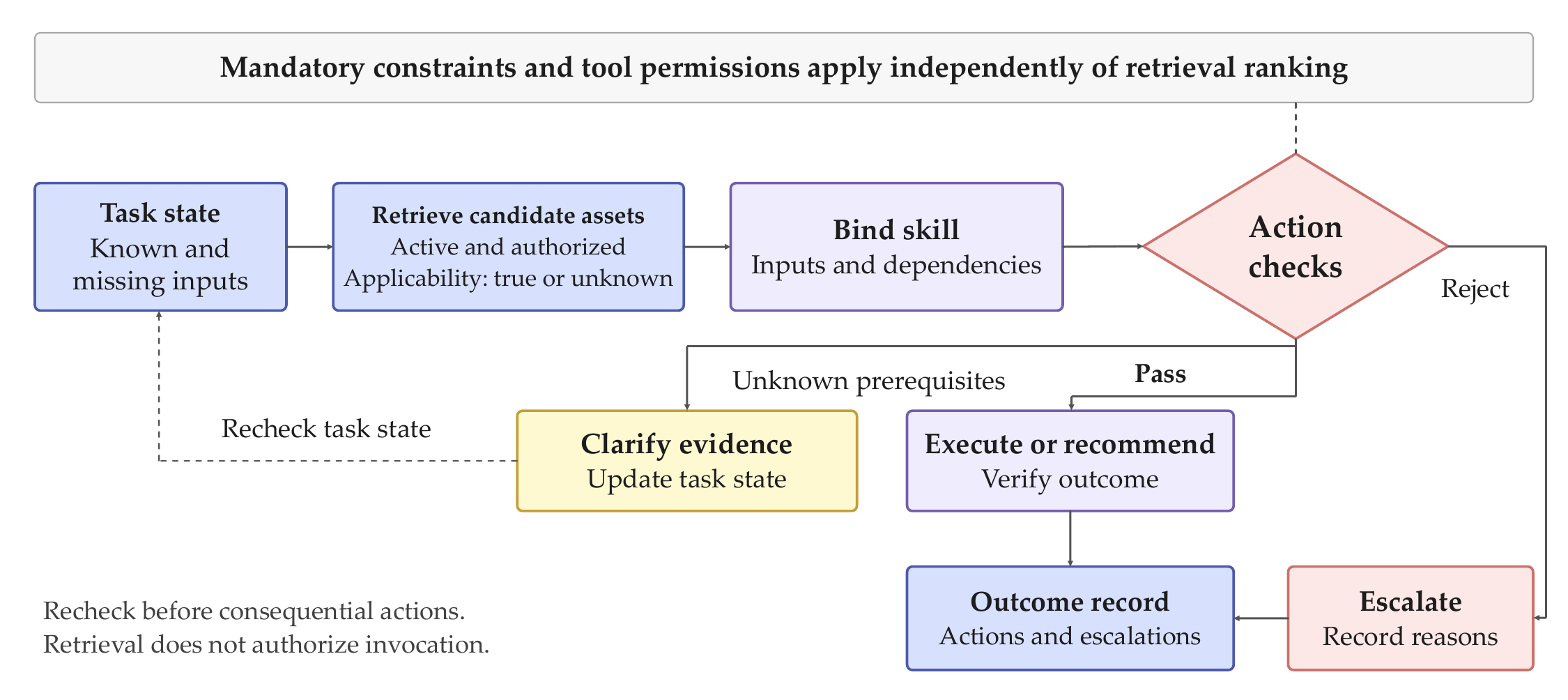}{Skill selection and execution. Retrieval retains active, authorized assets with true or unknown applicability. Invocation requires satisfied prerequisites and action checks. Missing information prompts clarification and another state check; a rejected action may lead to escalation. Actions and escalations are logged. Mandatory constraints apply independently of retrieval ranking.}{fig:runtime}

A skill organizes assets produced through nine-layer cognitive corpus construction into a procedure an agent can call. Its specification includes applicability, inputs, output structure, ordered steps, tool dependencies, constraints, stopping conditions, evidence references, and versions. Together, these fields make the procedure's inputs explicit and its execution and outputs checkable.

Skill construction starts from reviewed assets. Rules support judgments and state transitions. Best practices supply candidate procedures. Constraints and negative examples define checks. Corner cases identify when to pause or adjust the usual procedure. Each skill combines these elements into a complete procedure, with suggested wording where useful. A dependency list links each part of the procedure to its supporting assets, so corrections can trigger targeted revalidation.

The system retrieves and calls external assets during inference. Rules, constraints, and skills can therefore be updated without retraining the base model. Regression evaluation should check that a correction takes effect and that unrelated tasks do not degrade.

\subsection{Selection and execution}
Let $\asset$ be the asset set, $\sigma(r)$ an asset's status, $\phi_r(x)$ its three-valued applicability after considering prerequisites and exclusions, and $\mathrm{access}(r,x)$ its access permission. For task state $x$, retrieval keeps active, authorized assets whose applicability is true or unknown. Invocation uses a stricter subset:
\begin{align}
\mathcal{R}_x &= \{r\in\asset:\sigma(r)=\mathrm{active},\ \mathrm{access}(r,x)=\mathrm{allowed},\ \phi_r(x)\ne\mathrm{false}\},\\
\mathcal{X}_x &= \{r\in\mathcal{R}_x:\phi_r(x)=\mathrm{true}\}.
\end{align}
Retrieval and relevance ranking operate within $\mathcal{R}_x$. A relevant asset with unknown prerequisites can prompt a targeted question or a request for an observation. Only assets in $\mathcal{X}_x$ are eligible for invocation, which still requires dependency and action checks. False applicability excludes an asset. Applicability uses strong three-valued logic: a missing value is unknown, conjunction is false if any operand is false and true only if all are true, and negating unknown remains unknown. Unknown cannot be treated as either false or satisfied. Ranking may consider relevance, evidence, and recency, and its policy should be included in the evaluation record.

The agent binds known values to skill inputs, identifies missing prerequisites, and proposes a next step (Figure~\ref{fig:runtime}). Before an action with material consequences, it checks constraints and tool permissions. Mandatory checks operate independently of relevance ranking and top-$k$ truncation. An unknown mandatory condition blocks the action. Clarification targets fields whose values could change the next decision, and the updated state is checked again. Tool outputs are observations, not instructions that can override the task or platform controls. Output validation checks the required schema, reported evidence, and stopping conditions. A failed check can lead to a bounded retry, more information, a revised plan, or human referral.

Execution need not follow a fixed chain. Skills can branch, but branch conditions and allowed actions must be checkable. A new observation can invalidate a previously satisfied prerequisite, so checks must occur before action rather than only at task entry. The task record preserves the actual sequence, including failed attempts and human intervention.

Runtime logs record asset and skill calls for tracing and branch testing. The next section examines how experience processed by nine-layer cognitive corpus construction supports judgments, boundary checks, and procedures in actual runs.

\section{Case Study and Task Runs}
\label{sec:case}
\subsection{Comparing the three configurations in workplace injury mediation}
To examine the use of extracted experience in mediation, we compared the base model, raw-corpus RAG, and \product configurations for a workplace injury mediator. This practitioner supplied just 2 interview files, which produced 167 experience assets, and completed rule-library cross-review jobs.

One representative task asked whether a traffic accident during a detour to collect a child after work could qualify as a workplace injury. The answer also had to identify evidence about the route, timing, and purpose, and set an order for addressing disputed issues in mediation. The base model provided a general framework and evidence checklist but cited no specific regulation and did not clearly distinguish mediation from formal injury determination. Raw-corpus RAG cited the \emph{Regulations on Work-Related Injury Insurance} and proposed mediation steps. The \product agent additionally called the ``Mediation Request Assessment and Focus'' skill and consulted rule and corner-case libraries. It distinguished the boundaries of injury determination, mediation authority, and high-risk issues, and generated an output file. It was rated best on all 8 comparison items, including 1 tie with raw-corpus RAG.

\begin{table}[htbp]
\centering
\caption{Three-configuration comparison for workplace injury mediation. All configurations assess an accident during a detour to collect a child after work, the evidence needed, and the order of mediation steps.}
\label{tab:mediation_comparison}
\begin{tabularx}{\linewidth}{@{}>{\raggedright\arraybackslash}p{0.13\linewidth}XXX@{}}
\toprule \lsftablehead{Item} & \lsftablehead{A: Base model} & \lsftablehead{B: Raw-corpus RAG} & \lsftablehead{C: \product}\\\midrule
Approach & General analysis and an evidence checklist. & Cites work injury insurance regulations and proposes mediation steps. & Calls the mediation assessment skill and combines rules with corner cases.\\
Boundaries & Does not clearly separate mediation from formal injury determination. & Identifies some procedural and risk boundaries. & States that mediation does not replace formal determination and distinguishes accident scenarios and risks.\\
Output & General principles and evidence suggestions. & Legal grounds and process advice. & A structured process and an output file.\\
Comparison & Rated lowest on all 8 items. & Ties with C on 1 item. & Rated best or tied for best on all 8 items.\\
\bottomrule
\end{tabularx}
\end{table}

A second recorded question concerned a work-related injury with grade-ten disability, part-time employment, and several compensation amounts. The base model focused on legal principles and inferring amounts. Raw-corpus RAG added practical calculations and mediation advice. The \product agent called a mediation skill and used rules and best practices to organize wage evidence, compensation calculations, reasons for employment termination, and judicial confirmation into a complete process. It also generated a ``Preliminary Analysis of a Workplace Injury Mediation Case'' file.

On this practitioner's 8 evaluation questions, A, B, and C scored 73.34, 80.80, and 90.58, respectively. C exceeded B by 9.78 points. Mean end-to-end times were 13, 31, and 53 seconds. Structured experience improves the completeness and usefulness of professional analysis by supplying task boundaries, judgment conditions, and executable procedures. The comparison shows the added value of organizing source knowledge into experience assets and skills. The next two sections describe the evaluation setup and results across domains.

\section{Evaluation Design}
\label{sec:evaluation}
We systematically compare three configurations on a common evaluation protocol to assess the effectiveness of nine-layer cognitive corpus construction and use run records to explain how assets and skills contribute to task performance. The evaluation uses authorized samples from 20 randomly selected practitioners and covers 177 questions and 531 responses. A uses the base model, B retrieves the raw corpus, and C uses \product experience assets and skills.

\reportfigure{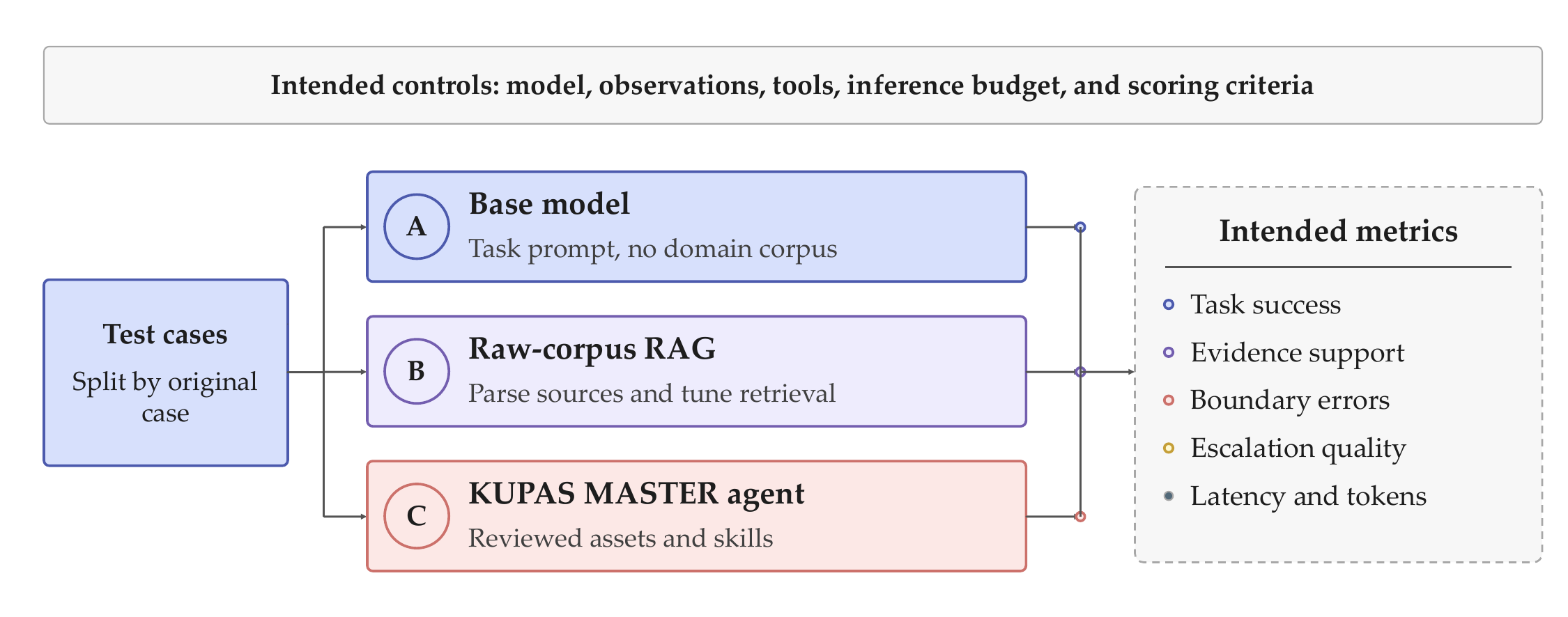}{The three-configuration comparison. Direct inference provides a model-only reference. Raw-corpus RAG measures the benefit of access to source records. The \product agent evaluates the complete experience-guided configuration. All three answer the same tasks under common input requirements and scoring criteria.}{fig:evaluation}

\subsection{Evaluation objectives}
The evaluation measures the \product agent's improvement over the base model and raw-corpus RAG. We analyze the gains alongside case representation, experience collection, consolidation, skill construction, and boundary controls to connect task results with the platform workflow.

Tasks should have clear inputs, bounded goals, and assessable outputs. AgentBench and benchmarks for web, enterprise, desktop, and software tasks illustrate environment-based evaluation~\citep{liu2024agentbench,zhou2024webarena,drouin2024workarena,xie2024osworld,jimenez2024swebench}. Benchmarks for tool and user interaction and scientific workflows further examine dialogue consistency and output validation~\citep{yao2025taubench,barres2025tau2,scienceagent2025}. Mediation tasks can test whether an agent identifies missing facts, selects the next information-gathering step, checks a response against reviewed criteria, or chooses an escalation path. Offline evaluation measures the accuracy and usefulness of advice. Evaluation in actual use can additionally track outcomes such as dispute resolution. Studies of professional writing and human-AI collaboration suggest focusing on task outcomes and comparing against the stronger of human-only and AI-only performance~\citep{noy2023productivity,vaccaro2024combinations}.

\subsection{Design principles}
Figure~\ref{fig:evaluation} shows the comparison. A receives the base model, task prompt, and observations available at task entry, with no domain retrieval. B adds retrieval from raw domain records, with the parsing and indexing needed to support it. C uses assets and skills constructed from those records. Across the experiments, each configuration answers independently under the same tasks, input requirements, and scoring rubric. Model settings, sampling parameters, tool permissions, and runtime resources follow a common protocol, as do the rules for providing task clarification. This consistent setup makes the results directly comparable.

B and C draw on the same authorized source materials, including practitioner interviews. B retrieves the original content, while C uses experience assets and skills derived from it. This shared information base allows the comparison to measure the value added by experience structuring and skill construction. Retrieval, skill calls, and end-to-end time are retained to analyze quality and resource use. Configuration records cover B's parsing, chunking, indexing, and retrieval, and C's corpus snapshot, experience retrieval, and skill versions.

\subsection{Experimental setup and data}
\paragraph{Sample processing.}
\product is commercially deployed and has accumulated experience from many practitioners. This report uses a random sample of 20 authorized users. Their 1,576 source files yield 30,762 retrievable chunks, 23,024 individual experience records, and 13,113 organizational assets after consolidation. The corresponding Elasticsearch indices contain 43,882 documents.

Shared identifiers link processing stages. The file inventory records sources, formats, and review status. Individual records link to contributors and cases. Organizational assets add types, versions, and source relationships. Index documents support retrieval and agent use. The sources span 13 formats, and 17 of the 20 practitioners have all six asset types. These records support analysis of corpus composition, asset distributions, and use across tasks.

\paragraph{Tasks and run configurations.}
The 177 evaluation questions each receive an independent answer from A, B, and C, producing 531 responses. Domains include finance and accounting, community governance, engineering quality, safety oversight, healthcare, mediation, and emergency management. Table~\ref{tab:evaluation_setup} summarizes the setup. The comparison evaluates experience use during inference: raw-corpus retrieval and structured experience assets improve task performance without retraining the base model. Reviewed experience also provides reusable material for subsequent model adaptation.

\begin{table}[!t]
\centering
\caption{Experimental setup and data composition.}
\label{tab:evaluation_setup}
\begin{tabularx}{\linewidth}{@{}>{\raggedright\arraybackslash}p{0.20\linewidth}X@{}}
\toprule \lsftablehead{Item} & \lsftablehead{Configuration and scale}\\\midrule
Sample & 20 randomly selected authorized practitioners, covering finance and accounting, community governance, engineering quality, safety oversight, healthcare, mediation, and emergency management.\\
Tasks & 177 questions, independently answered by A, B, and C, for 531 responses.\\
Difficulty & 8 easy, 83 medium, and 86 hard questions.\\
Configurations & A: base model; B: raw-corpus RAG; C: \product agent.\\
Scoring & A shared seven-dimension rubric applied to independent answers to the same tasks.\\
Run traces & Individual records for 5 practitioners and 53 questions, covering retrieval, library calls, and skill execution.\\
\bottomrule
\end{tabularx}
\end{table}

\paragraph{Scoring and run records.}
An LLM applies a shared rubric to score A, B, and C on the same questions across seven dimensions: result correctness, output actionability, depth of professional judgment, evidence sufficiency and accuracy, appropriate tool and skill use, accuracy in understanding requirements, and boundaries and compliance. Their weights are 22, 15, 15, 15, 11, 11, and 11, summing to 100. Main-text score means average the platform-reported practitioner summaries equally. These summaries are rounded at source; question-level distributions and bootstrap intervals are recomputed from the weighted dimension scores. Small differences in the last decimal can arise from that rounding. Displayed aggregates use decimal round-half-up at the stated precision.

The evaluation results support paired and stratified analysis by practitioner, domain, difficulty, and task form. Detailed run traces are available for 5 practitioners and 53 questions, showing knowledge-base retrieval, raw-corpus access, asset-library calls, and skill execution. In C, 52 runs retrieved from the knowledge base and 46 called skills. Rules, corner cases, best practices, constraints, and negative examples all appear in the traces. These records connect final scores to actual asset use.

Evaluation materials, run outputs, and summaries share a common data structure. Practitioner summaries retain question counts, configuration scores, failed questions, end-to-end time, and dimension scores alongside question-level results. Task records remain linked to the assets derived from their source materials. In workplace injury mediation, for example, the agent calls the ``Mediation Request Assessment and Focus'' skill and combines rules and corner cases to address procedural boundaries, fact verification, and risks. The dataset thus supports both outcome comparisons and analysis of the path from materials to assets, skill calls, and task outputs.

\subsection{Metrics and judgment}
Task performance is measured by a \emph{composite score}, the weighted sum of the seven dimension scores, and a \emph{score pass rate}, the fraction of questions scoring at least 60 points. The framework also defines task success for subsequent workflow evaluations as meeting a predefined domain rubric and all mandatory boundary conditions. Each run then has a binary outcome. For $K$ prespecified stochastic runs, let $s_{im}=K^{-1}\sum_{k=1}^{K}s_{imk}$ be the mean success rate of method $m$ on case $i$, with $s_{imk}\in\{0,1\}$. Its paired difference from baseline $b$ is
\begin{equation}
\widehat{\Delta}_{m,b}=\frac{1}{N}\sum_{i=1}^{N}(s_{im}-s_{ib}).
\end{equation}
For these workflow evaluations, the protocol fixes the rubric, mandatory failure conditions, repeat count, and stratum weights before testing, and reports stratum results, macro-averages, and deployment-weighted scores. Each original case is the statistical unit, with all its prespecified runs included in the case mean. Case-level resampling estimates intervals for paired differences while preserving within-case dependence. For the score gains reported here, practitioner-cluster resampling retains dependence among a practitioner's questions and produces 95\% intervals.

Secondary measures include factual support, procedural completeness, applicability errors, prohibited actions, unnecessary refusals, and escalation quality. Evidence faithfulness measures how well a cited passage supports the corresponding claim. Prior work studies claim-level factuality and citation verification~\citep{min2023factscore,liu2023verifiability}; RAGAs and RAGChecker distinguish retrieval quality from generation quality~\citep{es2024ragas,ru2024ragchecker}. The evaluation framework distinguishes citation coverage from factual support, checks claims against sources, and measures latency and token use under a common runtime budget. Collection and expert review costs form a separate part of the cost analysis.

Extensions to human evaluation will use randomized output order, masked system identities where feasible, a fixed rubric, and independent review of a subset, with reviewer qualifications, agreement, and adjudication recorded. Human checks complement model scoring for calibration and scale~\citep{zheng2023judge}. Workflow evaluations will connect these assessments to business outcomes over a defined observation period.

\section{Experimental Results and Analysis}
Table~\ref{tab:main-results} shows gains from both raw-corpus RAG and the \product agent. Calculated before rounding the aggregate means, B improves on A by 9.13 points, and C improves on B by 9.83 points. C gains across all scoring dimensions, including 11.1 points in evidence sufficiency and accuracy and 9.0 in boundaries and compliance (Figure~\ref{fig:dimdelta}). The 95\% practitioner-cluster bootstrap interval for C minus B is [9.20, 10.45], confirming a consistently positive score gain.

\begin{table}[htbp]
\caption{Overall performance of the three configurations. Higher scores and pass rates are better; lower latency is better. Scores and latency weight practitioners equally; pass rates weight the 177 questions equally, with a pass threshold of 60.}
\label{tab:main-results}
\begin{tabularx}{\linewidth}{@{}X>{\raggedright\arraybackslash}p{0.105\linewidth}>{\raggedright\arraybackslash}p{0.105\linewidth}>{\raggedright\arraybackslash}p{0.13\linewidth}>{\raggedright\arraybackslash}p{0.16\linewidth}>{\raggedright\arraybackslash}p{0.09\linewidth}@{}}
\toprule
\lsftablehead{Method} & \lsftablehead{Composite score} & \lsftablehead{Score pass rate (\%)} & \lsftablehead{Evidence sufficiency and accuracy} & \lsftablehead{Boundaries and compliance} & \lsftablehead{Time (s)}\\\midrule
A: Base model & 70.63 & 87.57 & 62.18 & 74.52 & 16.2\\
B: Raw-corpus RAG & 79.75 & 98.87 & 77.27 & 79.19 & 34.0\\
C: \product & 89.58 & 100.00 & 88.32 & 88.23 & 57.1\\
\bottomrule
\end{tabularx}
\end{table}

\subsection{Composite scores}
\begin{figure}[!t]
\centering
\includegraphics[width=0.96\linewidth]{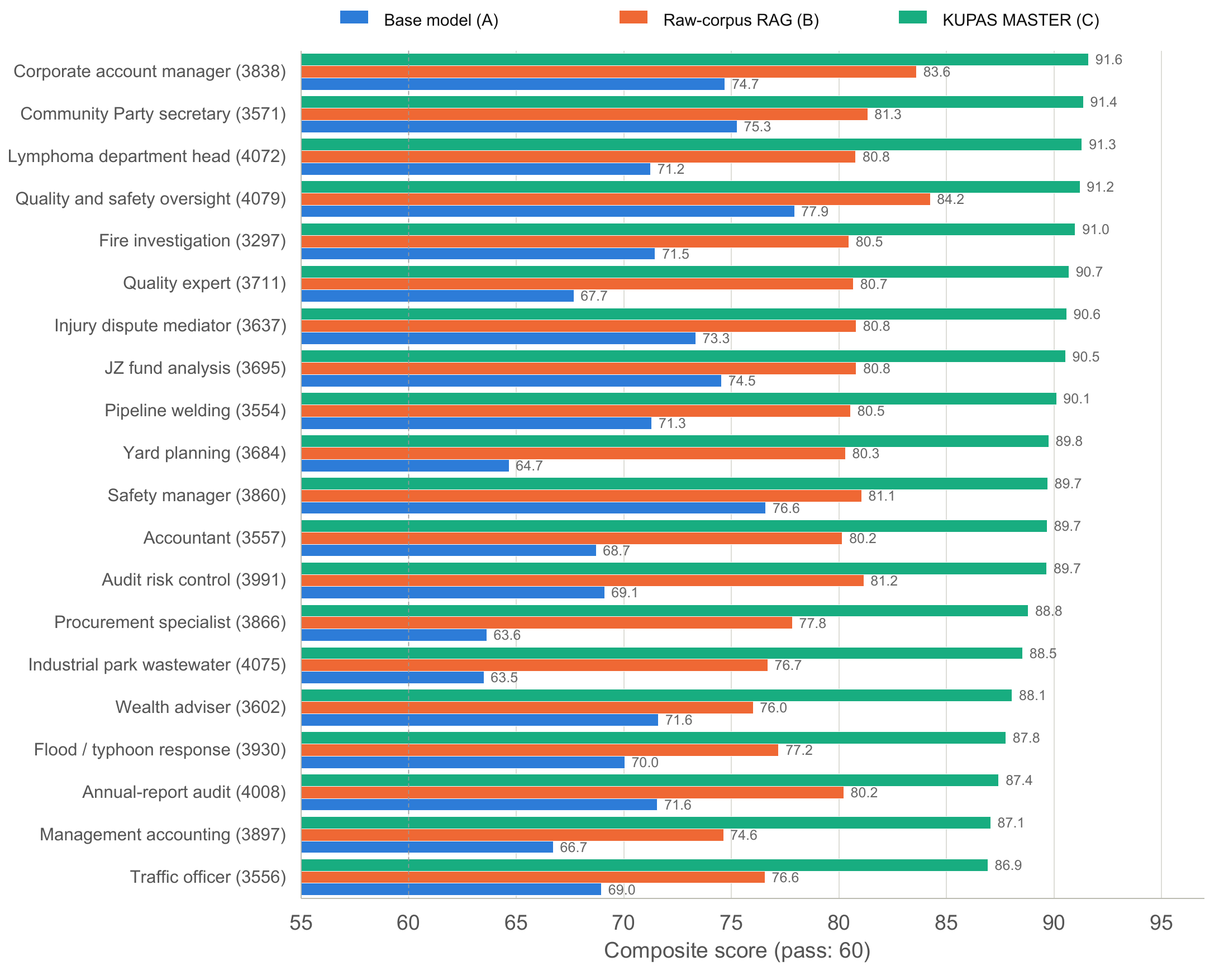}
\caption{Composite scores for the 20 practitioners. Equal weighting across practitioners gives means of 70.63, 79.75, and 89.58 for A, B, and C, with $C-B=+9.83$ and $C-A=+18.96$. C exceeds both baselines for every practitioner.}
\label{fig:overall}
\end{figure}
Figure~\ref{fig:overall} shows the composite score for each practitioner. The mean rises from 70.63 for A to 79.75 for B and 89.58 for C. Differences computed before rounding the aggregate means are 9.13 points for B minus A, 9.83 for C minus B, and 18.96 for C minus A. All 20 practitioner means follow $A<B<C$. At the question level, C exceeds both A and B on 176 of 177 questions, demonstrating consistent improvement across professional tasks.

\subsection{Performance by dimension}
\begin{figure}[htbp]
\centering
\includegraphics[width=\linewidth]{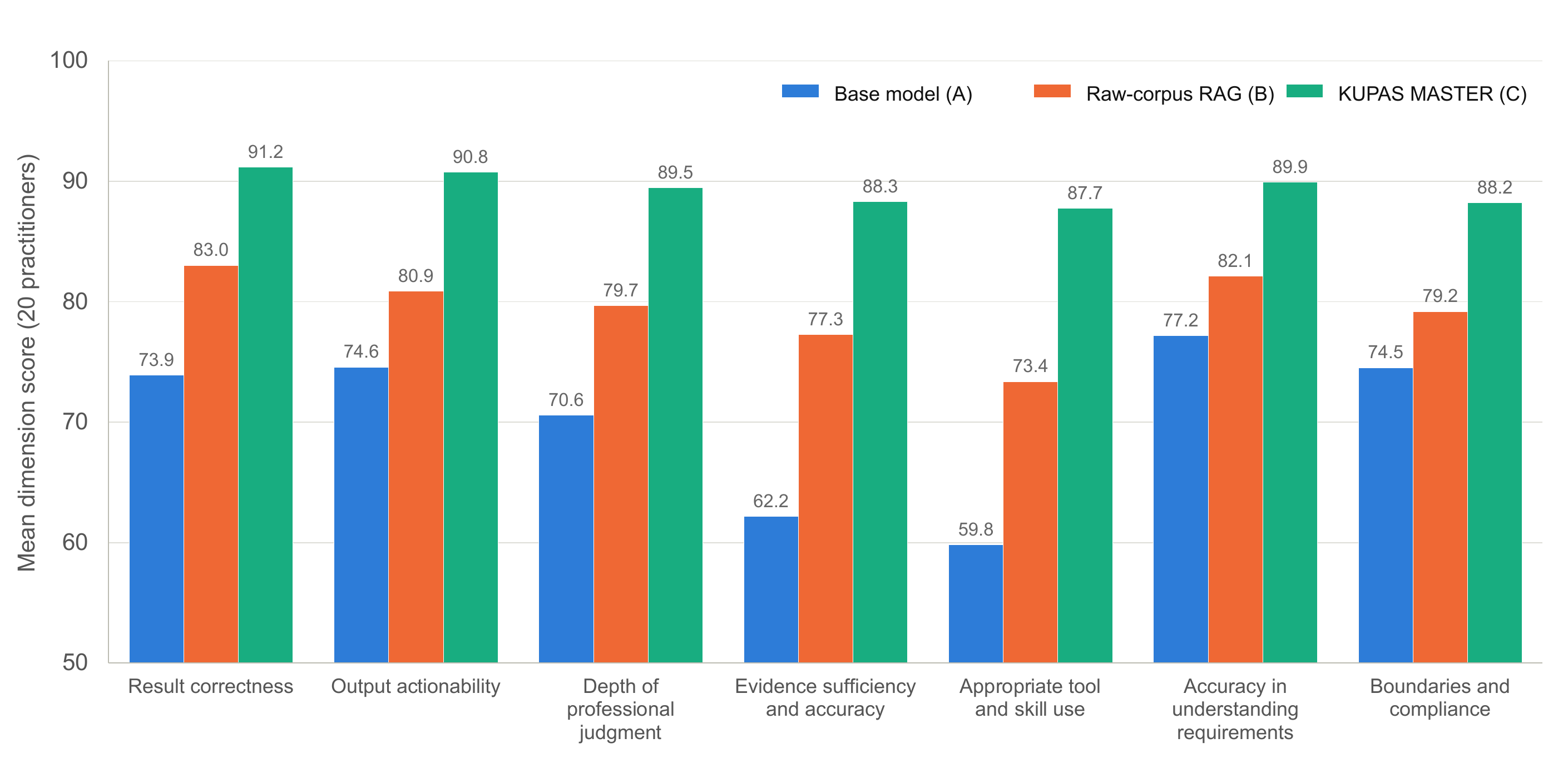}
\caption{Mean scores in seven dimensions, weighting the 20 practitioners equally. C scores range from 87.7 to 91.2. For every practitioner, C exceeds B in every dimension.}
\label{fig:dimension}
\end{figure}
Figure~\ref{fig:dimension} compares result correctness, output actionability, depth of professional judgment, evidence sufficiency and accuracy, appropriate tool and skill use, accuracy in understanding requirements, and boundaries and compliance. C's mean scores range from 87.7 to 91.2, and all 20 practitioners score higher in C than B on every dimension. The gains therefore cover output quality, evidence, judgment, execution, and compliance boundaries.

\subsection{Gains at each stage}
\begin{figure}[htbp]
\centering
\includegraphics[width=\linewidth]{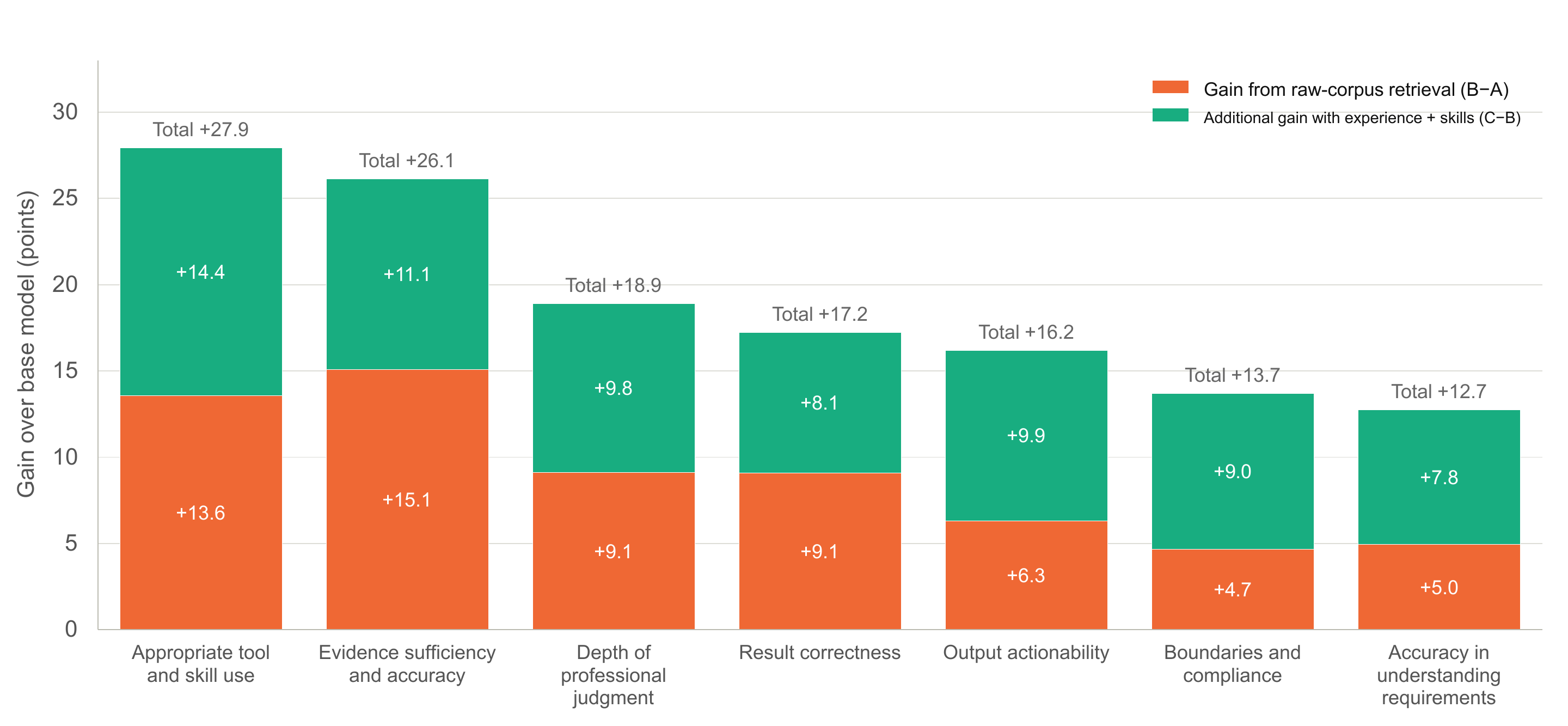}
\caption{Score gains by dimension. Orange shows the gain from raw-corpus retrieval ($B-A$); green shows the additional gain from experience assets and skills ($C-B$). Labels above the bars give the total ($C-A$). All dimensions weight the 20 practitioners equally.}
\label{fig:dimdelta}
\end{figure}

Figure~\ref{fig:dimdelta} separates the change from A to C into two stages. B minus A measures the gain from raw-corpus RAG. C minus B measures the additional gain from assets and skills constructed through the nine-layer approach. The overall gains, 9.13 and 9.83 points, are similar in size. Both access to source records and the structured experience configuration improve scores under the shared evaluation process.

The practitioner means, seven dimensions, and staged comparisons all favor C. The next analysis uses asset calls, skills, and run traces to examine how the experience configuration participates in task execution.

\subsection{Sources of gains and runtime behavior}
We examine organizational asset counts, run logs, and representative tasks to understand C's gains over raw-corpus RAG. The analysis considers asset organization, retrieval, skill use, source formats, and corpus size.

The 20 practitioners contributed 23,024 published individual entries, which became 13,113 organizational assets after consolidation. The process combines duplicate or similar experience and organizes individual entries into shared assets. Consolidation adapts to each practitioner's source material, with the reduction reflecting differences in repetition, content structure, and organization. Retrieval then operates over structured rules, best practices, constraints, negative examples, and corner cases.

For the 5 practitioners and 53 questions with complete individual traces, C performs knowledge-base retrieval in 52 runs and calls skills in 46. Rules, corner cases, best practices, constraints, and negative examples appear in 46, 35, 33, 28, and 18 runs, respectively. Compared with B's main reliance on raw-corpus retrieval, C combines several asset types and calls skills where needed. Table~\ref{tab:asset_mechanism} summarizes their observed roles.

\begin{table}[!t]
\centering
\caption{Asset and skill use in 53 C runs for 53 questions. Counts are by run and can overlap.}
\label{tab:asset_mechanism}
\begin{tabularx}{\linewidth}{@{}>{\raggedright\arraybackslash}p{0.17\linewidth}p{0.09\linewidth}X@{}}
\toprule \lsftablehead{Asset type} & \lsftablehead{Runs} & \lsftablehead{Observed role and example tasks}\\\midrule
Rules & 46 & Express judgment conditions as rules, thresholds, and requirements, including flood dispatch decisions and engineering quality escalation criteria.\\
Best practices & 33 & Supply established procedures and steps that turn experience into a task process.\\
Corner cases & 35 & Add exceptions, high-risk situations, and business boundaries that routine rules may miss.\\
Constraints & 28 & Identify prohibited actions or required prerequisites for procedural and compliance checks.\\
Negative examples & 18 & Provide recurring errors and inappropriate responses to help recognize risky decisions and exceptions.\\
Skills & 46 & Organize retrieved experience into steps, checks, responsible roles, and structured deliverables.\\
\bottomrule
\end{tabularx}
\end{table}

Examples make these roles concrete. In flood and typhoon response, rules, best practices, and corner cases support judgments about water levels, dispatch order, and de-escalation conditions. In safety management, negative examples and corner cases flag hazards such as unplanned rescue attempts in confined spaces and distinctions between ordinary hazards and major accident risks. In social insurance auditing, skills organize experience into plans, checklists, review forms, and ledgers. Assets thus contribute conditions, risk boundaries, and procedures as well as retrieved text.

The gains span appropriate tool and skill use, evidence sufficiency and accuracy, depth of professional judgment, output actionability, and boundaries and compliance. Removing the appropriate tool and skill use dimension and renormalizing the remaining weights to 100 leaves an average C-minus-B gap of about 9.3 points, close to the full-score gap of 9.83. The difference extends to evidence use, judgment, and task outputs.

All 20 practitioners gain across different corpus sizes and formats. Raw chunk counts, organizational asset counts, and library sizes show no stable relationship with C's gain over B. Spreadsheets can produce many cell fragments, JSON contains structural fields, and documents usually provide more continuous text. The platform organizes these sources into a common representation. The consistent gains across these varied corpora demonstrate the practical value of organizing heterogeneous experience for task use.

The traces show a connected process: experience extraction produces assets, retrieval combines relevant assets for the current question, and skills organize them into steps, checks, and deliverables. This workflow turns experience into stronger evidence, clearer judgments, actionable outputs, and explicit boundary checks, connecting the score improvements to concrete task behavior.

\subsection{Efficiency}
Across 20 practitioners and 177 questions, mean end-to-end latency, weighting practitioners equally, is 16.2 s for A, 34.0 s for B, and 57.1 s for C. The means of practitioner-level 90th-percentile (P90) latencies are 21.1, 41.1, and 68.4 s. C's mean latency is about 3.5 times A's and 1.7 times B's.

In the detailed-trace subset of 5 practitioners and 53 questions, C performs knowledge-base retrieval in 52 runs and calls skills in 46. Library usage is 46 for rules, 35 for corner cases, 33 for best practices, 28 for constraints, and 18 for negative examples. Mean latency on this same subset is 17.4, 36.2, and 59.1 s for A, B, and C.

The platform delivers stronger professional judgments, better supporting evidence, and actionable outputs in about a minute on average. This response time accommodates experience retrieval, asset checks, and skill execution within a practical workflow for business analysis, planning, and professional review. The result is a substantial quality improvement with a response window suited to these tasks.

\section{Commercial Deployment}
\label{sec:deployment}
\product is a commercial platform with cloud and local deployment options. It separates corpus production, asset management, agent execution, and evaluation records, and uses permissions, versions, and run records to manage access across these stages. The experiments use the release dated September 16, 2026. The platform continues to evolve.

\subsection{Customized local deployment}
The platform supports deployment of corpus processing, asset management, retrieval, and agent execution on customer-owned servers or a private cloud. Domain terms, library structures, review rules, and skill workflows can be configured for an industry's requirements and business processes.

For applications with strict confidentiality requirements, local model inference and internal tool services can keep source materials, assets, task inputs, outputs, and logs within the customer environment. In this configuration, business data need not pass through external model APIs or third-party services. Authentication, role-based permissions, access isolation, and audit logs govern collection, sharing, and agent use. Customers can control where data are stored, how they are used, and who can access them, while retaining a record of operations.

\section{Related Work}
\label{sec:related}
\paragraph{Acquiring and organizing professional experience.}
The Critical Decision Method and Applied Cognitive Task Analysis use questions about specific incidents to recover cues, judgments, and strategies missing from routine records~\citep{klein1989critical,hoffman1998cdm,militello1998acta}. Organizational knowledge creation also studies how individual knowledge becomes shared practice~\citep{nonaka1994knowledge}. \product applies these ideas to the design of records for agents. Its nine dimensions organize expert judgments, actions, and feedback into experience that can be extracted, reviewed, and reused.

\paragraph{Retrieval and structured corpora.}
Dense Passage Retrieval and Fusion-in-Decoder study learned retrieval and generation from multiple passages~\citep{karpukhin2020dpr,izacard2021fid}. Self-RAG adds retrieval and self-critique decisions, while RAPTOR organizes recursive summaries~\citep{asai2024selfrag,sarthi2024raptor}. GraphRAG and LightRAG retrieve linked information through graphs~\citep{edge2024graphrag,guo2025lightrag}, and HippoRAG 2 studies nonparametric continual learning~\citep{gutierrez2025memory}. These approaches offer alternatives to a simple raw-corpus index. A longer context alone does not ensure that a model uses the relevant evidence~\citep{liu2024lost}. Our focus is the experience contributed by practitioners and the conditions under which it applies.

\paragraph{Memory and experience reuse.}
CoALA distinguishes memory and action components in language agents~\citep{sumers2024coala}. Generative Agents uses experience records and reflection to guide simulated behavior~\citep{park2023generative}. Reflexion and ExpeL derive reusable feedback from agent attempts~\citep{shinn2023reflexion,zhao2024expel}. A-Mem and Mem0 study structured or consolidated memory, while ACE and LangMem support evolving context and persistent memory~\citep{xu2025amem,chhikara2025mem0,zhang2025ace,langchain2025langmem}. \product focuses on externally collected practitioner accounts, claim-level evidence, conditional disagreements, and review before use. This approach brings practitioner expertise and its supporting evidence into agent memory and reuse.

\paragraph{Skills and execution frameworks.}
Toolformer studies learned tool use, and ReAct combines reasoning with environment interaction~\citep{schick2023toolformer,yao2023react}. Voyager maintains executable skills, and Agent Workflow Memory derives reusable workflows from past trajectories~\citep{wang2023voyager,wang2025workflow}. AutoGen, MetaGPT, DSPy, and AgentScope provide orchestration or program construction mechanisms~\citep{wu2024autogen,hong2024metagpt,khattab2024dspy,gao2025agentscope}. \product can supply reviewed content to these systems. Our focus is the evidence and dependency requirements of the procedures passed to an executor, building on existing orchestration work.

\paragraph{AI in professional and scientific work.}
AMIE studies conversational diagnosis, and subsequent work extends conversational AI to disease management~\citep{tu2025amie,lievin2026management}. Agents for scientific instruments and Co-Scientist study tool-based workflows and scientist-guided discovery, respectively~\citep{instrument2026,coscientist2026}. These studies inform evidence organization, expert assessment, and task validation in professional settings. The platform improves agents through external experience corpora accessed during inference, without requiring base-model retraining.

\section{Conclusion}
We presented \product, an experience engineering platform built around nine-layer cognitive corpus construction. It turns practitioners' tacit experience into structured assets that can be traced and reused. Six case elements, nine extraction dimensions, and six asset types connect case records to experience and callable skills. Semantic alignment, individual distillation, organizational consolidation, and cross-review retain sources, conditions, and disagreements. Explicit inputs, steps, dependencies, and stopping conditions connect these assets to task execution and evaluation feedback.

Using authorized samples from 20 randomly selected practitioners, the platform processed 1,576 source files into 23,024 individual records and 13,113 organizational assets. Across 177 questions and 531 responses produced under a common task protocol and evaluated with a shared rubric, the base model, raw-corpus RAG, and \product agent scored 70.63, 79.75, and 89.58 when practitioner means were weighted equally. The \product agent improved on raw-corpus RAG by 9.83 points and gained in all seven dimensions. Run records confirm that assets and skills supplied judgment conditions, identified risk boundaries, and organized task procedures. These results demonstrate the effectiveness of nine-layer cognitive corpus construction, with consistent advantages in task quality, evidence use, and boundary handling. The platform turns individual tacit experience into organizational knowledge and agent capabilities through a complete engineering workflow, from experience capture to practical application.

By bringing expert judgment, practical methods, and execution boundaries into agent workflows, \product provides a foundation for staff development, business collaboration, and professional services. Its traceable assets preserve organizational know-how and make professional experience reusable across tasks. Future development will expand the asset base and applications, streamline experience capture and reuse, and strengthen sharing, review, updates, and application feedback. These advances will help organizations build lasting knowledge assets and apply AI to increasingly demanding decisions and tasks.

\label{page:main-end}
\clearpage
\bibliographystyle{unsrtnat}
{\raggedright
\bibliography{refs}
}
\clearpage
\appendix
\section{Team Members}
\label{app:team}
The project and research teams of KUPAS MASTER are listed below.

\subsection{Project Leaders}
\begin{tabularx}{\linewidth}{@{}>{\columncolor{white}[0pt][\tabcolsep]}p{0.18\linewidth}p{0.18\linewidth}>{\columncolor{white}[\tabcolsep][0pt]}X@{}}
\toprule \lsftablehead{Name} & \lsftablehead{Affiliation} & \lsftablehead{Title} \\\midrule
Changmian Wang & KUPAS & CTO\\
Yuchao Ma & KUPAS & Director of Innovative Products; Technical Architect; Senior FDE\\
\bottomrule
\end{tabularx}

\subsection{Project Members}
\begin{tabularx}{\linewidth}{@{}>{\columncolor{white}[0pt][\tabcolsep]}p{0.18\linewidth}p{0.18\linewidth}>{\columncolor{white}[\tabcolsep][0pt]}X@{}}
\toprule \lsftablehead{Name} & \lsftablehead{Affiliation} & \lsftablehead{Title} \\\midrule
Xuchao Lu & KUPAS & Algorithm Expert; Senior FDE\\
Chen Zhang & KUPAS & AI Product Manager; Senior FDE\\
Ping Sun & KUPAS & AI Engineer; Senior FDE\\
Jiazheng Wang & KUPAS & AI Evaluation Expert; Senior FDE\\
Shan Wang & KUPAS & Senior FDE\\
\bottomrule
\end{tabularx}

\subsection{Research Leaders}
\begin{tabularx}{\linewidth}{@{}>{\columncolor{white}[0pt][\tabcolsep]}p{0.24\linewidth}p{0.18\linewidth}>{\columncolor{white}[\tabcolsep][0pt]}X@{}}
\toprule \lsftablehead{Name} & \lsftablehead{Affiliation} & \lsftablehead{Title} \\\midrule
Qinghua Zheng & Tongji University & Party Secretary of Tongji University; Academician of the Chinese Academy of Engineering\\
Xian-Sheng Hua & Tongji University & Executive Dean, Institute of AI for Engineering; Tenured Distinguished Professor\\
\bottomrule
\end{tabularx}

\subsection{Research Members}
\begin{tabularx}{\linewidth}{@{}>{\columncolor{white}[0pt][\tabcolsep]}p{0.18\linewidth}p{0.18\linewidth}>{\columncolor{white}[\tabcolsep][0pt]}X@{}}
\toprule \lsftablehead{Name} & \lsftablehead{Affiliation} & \lsftablehead{Title} \\\midrule
Hongzhi Li & Tongji University & Tenured Distinguished Professor; Former Principal Researcher and Chief Architect, Microsoft (USA)\\
Jianqiang Huang & Tongji University & Tenured Professor; Former Head of Alibaba City Brain\\
Kaihua Tang & Tongji University & Tenured Associate Professor; World's Top 2\% Scientists, 2024--2025\\
Ziqing Xia & Tongji University & Assistant Professor, Human Factors Expert\\
Ziyu Lu & Tongji University & PhD Student\\
Yihe Sun & Tongji University & Master's Student\\
Xuanwen Chen & Tongji University & Assistant Research Fellow\\
\bottomrule
\end{tabularx}

\clearpage
\section{Case Study: System Workflow}
\label{app:case-evidence}
\begin{figure}[H]
\centering
\includegraphics[width=\linewidth]{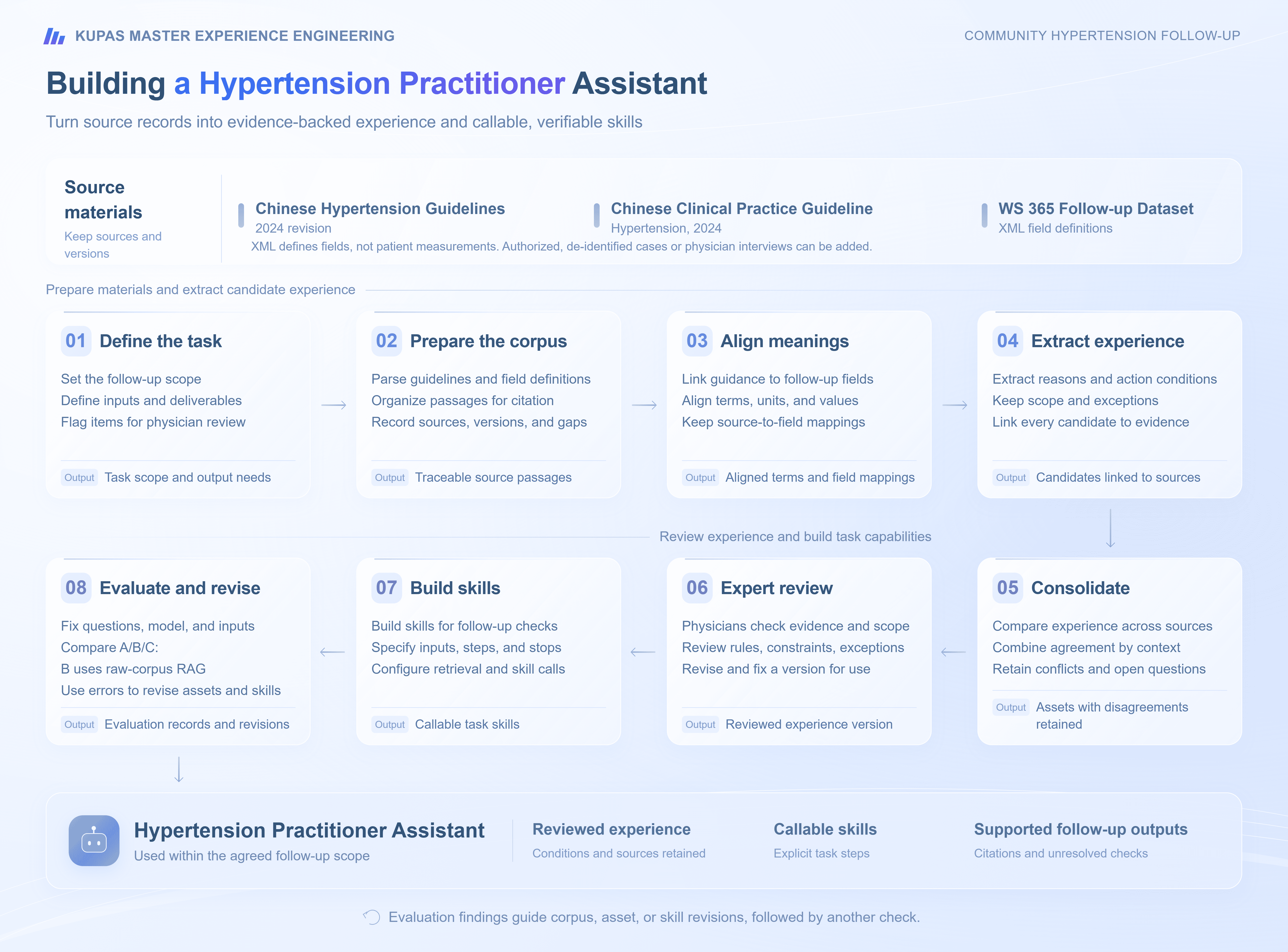}
\caption{Building, using, and evaluating a hypertension practitioner assistant.}
\label{fig:hypertension-workflow}
\end{figure}

Figure~\ref{fig:hypertension-workflow} illustrates how a hypertension practitioner assistant is built and used for community follow-up. The user first defines the task, inputs, and deliverables, imports guidelines and follow-up field definitions, and records sources, versions, and missing information. Semantic alignment connects the materials. Extraction then recovers judgment criteria, action conditions, and exceptions as candidates linked to evidence. Experience from different sources is consolidated and reviewed by domain physicians, then organized into skills with explicit inputs, steps, and stopping conditions. The agent uses these skills to produce follow-up outputs with citations and clear verification steps. Evaluation fixes the questions, model, and inputs, compares configurations, and feeds errors back into corpus, asset, or skill revisions. The XML in the figure provides follow-up field definitions for subsequent data entry.

\FloatBarrier

\section{Detailed Experimental Analysis}
\label{app:experiments}
This appendix brings together configuration scores, asset statistics, and run records to explain the platform's performance, experience organization, skill execution, boundary handling, domain coverage, and efficiency. Table~\ref{tab:experiment-evidence} summarizes ten aspects of the analysis and their supporting evidence. All configuration comparisons follow the common evaluation protocol described in Section~\ref{sec:evaluation}.

\begin{longtable}{@{}p{0.045\linewidth}>{\raggedright\arraybackslash}p{0.31\linewidth}>{\raggedright\arraybackslash}p{0.57\linewidth}@{}}
\caption{Ten aspects of platform performance and supporting evidence (20 practitioners, 19 platform industry labels, 177 questions, and 531 responses).}\label{tab:experiment-evidence}\\
\toprule \lsftablehead{ID} & \lsftablehead{Analysis focus} & \lsftablehead{Supporting evidence}\\\midrule
\endfirsthead
\toprule \lsftablehead{ID} & \lsftablehead{Analysis focus} & \lsftablehead{Supporting evidence}\\\midrule
\endhead
E1 & Overall improvement in task outcomes & Configuration scores, paired questions, resampling intervals, and below-threshold scores in the random authorized sample.\\
E2 & Added value from structuring the same information & Overall gain of the experience-and-skill configuration built from the same sources.\\
E3 & Useful, traceable experience representation & Asset fields, source relationships, automated checks, and separately recorded human review.\\
E4 & Consolidation into organizational assets & Individual-to-organizational asset counts, library versions, and consolidation records.\\
E5 & Skills in task execution & Skill calls, file generation, and task scores.\\
E6 & Boundary handling and reliability & Boundary score gains and associations between asset use and boundary-judgment rows.\\
E7 & Performance across corpus sizes & Performance across sample sizes and rank correlations for the two stages.\\
E8 & Traceable updates and repeated execution & Library versions, repeated-question reports, and retrieval records across runs.\\
E9 & Consistent performance across professional domains & Task performance in multiple professional settings, each using domain-specific corpora.\\
E10 & Runtime cost and efficiency & End-to-end latency and percentile statistics by practitioner and configuration.\\
\bottomrule
\end{longtable}

\noindent\textbf{Analysis and supporting evidence}
\begin{itemize}
\setlength{\itemsep}{4pt}
\item \textbf{E1: Overall task performance.}
The comparison uses authorized samples from 20 randomly selected practitioners, covering 177 questions and 531 responses. Scores rise from 70.63 for A to 79.75 for B and 89.58 for C. C gains 18.96 points over A and 9.83 over B, equivalent to relative increases of 26.8\% and 12.3\%. All practitioner means follow $A<B<C$. C wins 176/177 paired question comparisons against B and 177/177 against A. Counts of responses scoring below 60 fall from 22 to 2 to 0. The 95\% interval is $[9.20,10.45]$ for the practitioner-weighted gap of 9.83 and $[9.02,10.45]$ for the question-weighted gap of 9.72, using 100,000 cluster bootstrap samples with seed 20260926.

\item \textbf{E2: Structuring the same source information.}
The raw-corpus RAG and structured-asset configurations use the same uploaded materials. C adds 9.83 points (12.3\%) over B, with the same direction of change across all 20 practitioners and all 7 dimensions. The raw retrieval stage contributes 9.13 points over A. The shared-source comparison demonstrates the added value of turning raw information into structured experience and executable skills.

\item \textbf{E3: Representation and reliability.}
The six libraries contain 13,113 organizational assets. Their fields include triggers, exceptions, judgment logic, evidence spans, source links, and confidence, supporting checkable citations. Across the six library types, 37 cross-review jobs cover 818 candidates, of which 486 are marked as passed (59.4\%). The best-practice pass rate is 70\%, and the mean of the 37 job-level scores is 79.5. Human review status is recorded independently. The ratio of organizational assets to retrievable chunks ranges from 0.02 to 11.9, reflecting different source forms, from tables to large case collections.

\item \textbf{E4: Organizational consolidation.}
The platform converts 23,024 published individual entries into 13,113 organizational assets through consolidation of agreement, disambiguation, and context annotation. All six libraries are present for 17/20 practitioners. Organizational libraries are rebuilt with version identifiers and enter the shared index of 43,882 Elasticsearch documents. Entries retain a consolidation-method field, making each asset's processing history traceable.

\item \textbf{E5: Skills and task execution.}
Skills organize extracted assets into executable procedures. Among 217 three-configuration comparison reports, C calls skills in 166 (76.5\%) and generates files in 116 (53.5\%); A and B make no skill calls. In the 53 C runs with detailed logs, 46 call skills and 46 retrieve rules. C gains 14.38 points in appropriate tool and skill use and 9.89 in output actionability over B. The records show experience being used to produce deliverables as well as advice.

\item \textbf{E6: Boundary assets and reliability.}
The boundaries and compliance score increases by 9.03 points from B to C, about 1.9 times the 4.67-point gain from A to B. For 19 of 20 practitioners, the C-minus-B gain on this dimension exceeds the B-minus-A gain. Reports that use corner cases, constraints, or negative examples have 95 C-only wins in 99 boundary-judgment rows, compared with 94/127 (74.0\%) in reports without those calls. Boundary rows are identified by judgment-item names mentioning boundaries, risk, or applicability; library use is counted at the report level. Run records show how these assets supply concrete checks, such as verifying the grounds for judging a transfer unlawful and checking compliance evidence beyond proof of delivery.

\item \textbf{E7: Data scale and usefulness.}
Small, focused collections deliver strong performance. The community Party secretary has 7 files and 36 assets, with a composite score of 91.36, ranking second. The workplace injury mediator has 2 interviews and 167 assets, scoring 90.58. The Spearman correlation between asset count and the C-minus-B gain is only 0.07. Raw chunk count correlates with the B-minus-A gain at 0.61 ($p\approx0.004$), suggesting different relationships at the two stages. All 20 practitioners improve across corpora spanning two orders of magnitude.

\item \textbf{E8: Updates and version management.}
The libraries support batch rebuilding and retain library-level versions. Current versions for 13 practitioners share a timestamp prefix. The organizational libraries are updated as their corpora evolve. In 19 repeated runs of an accountant's equity-method investment income question, every comparison row includes C among the winners. A wealth-adviser question retrieves between 1 and 5 library types, including the raw corpus, across runs. Version records make experience updates traceable, while repeated runs document how the agent retrieves and applies experience across executions.

\item \textbf{E9: Consistent gains across industries.}
The sample covers 19 platform industry labels. All 20 practitioners improve in all 7 dimensions, and C exceeds B on 176/177 questions. Applications include firefighting, energy, traffic policing, finance and accounting, community governance, mediation, ports, economic crime investigation, rail transit, emergency management, social insurance funds, auditing, healthcare, environmental management, and construction. The platform consistently improves task performance across these professional settings by turning their varied source materials into usable domain experience.

\item \textbf{E10: Runtime cost and efficiency.}
Practitioner-weighted mean latencies are $16.2/34.0/57.1$ s for A/B/C. The corresponding means of practitioner P90 values are $21.1/41.1/68.4$ s. Ratios of unrounded mean times put C at $3.53\times$ A and $1.68\times$ B. C's practitioner means range from 43.6 to 78.3 s, and the mean of practitioner-level 95th-percentile (P95) latencies is 71.6 s. Latency is positively associated with the number of libraries retrieved, skill calls, and file generation. The platform combines these response times with higher accuracy and more complete outputs for professional analysis, planning, and review.
\end{itemize}

\section{Additional Results and Plots}
\label{app:illustrative}
This appendix visualizes the evaluation data in more detail. Score distributions and difficulty groups weight the 177 questions equally. Practitioner-level composite scores, dimension means, and mean latency weight the 20 practitioners equally. The captions identify the aggregation used.

\begin{figure}[htbp]
\centering
\includegraphics[width=\linewidth]{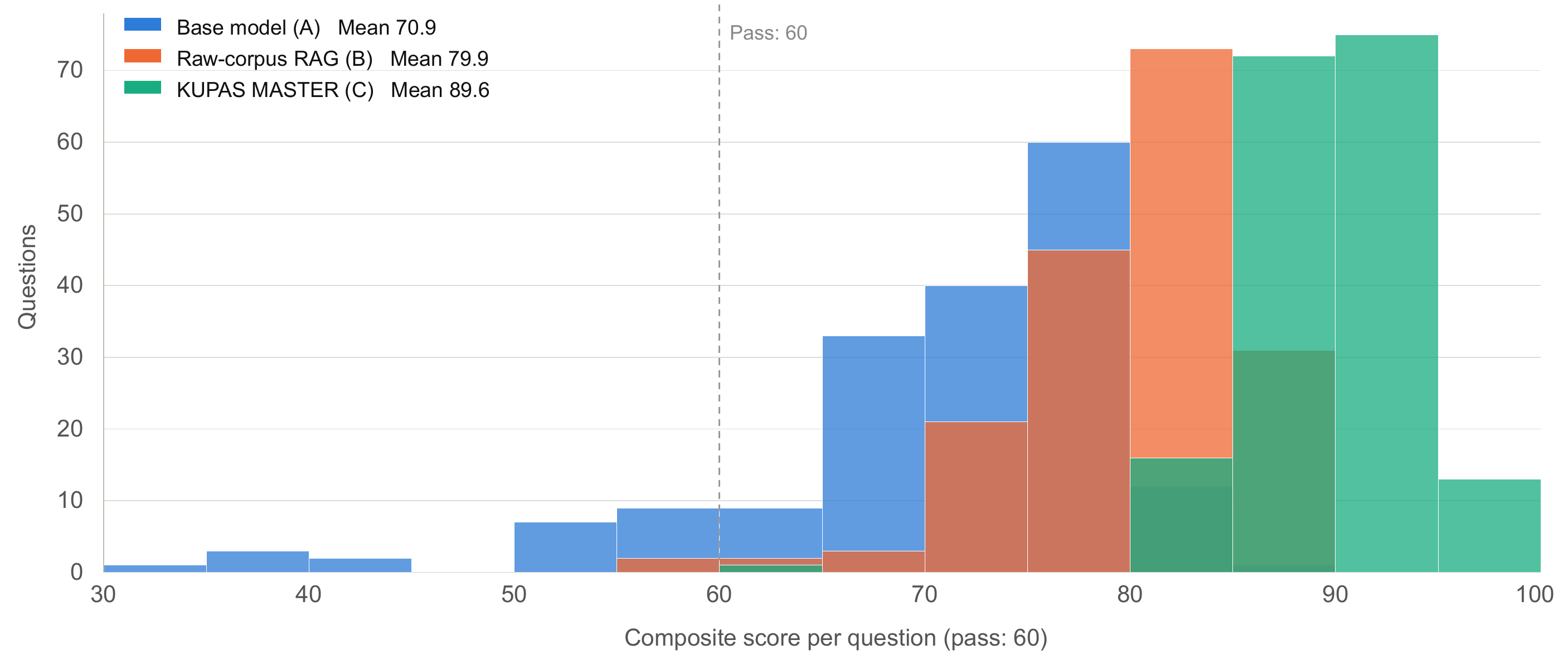}
\caption{Distribution of 531 scores for 177 questions, with 177 scores per configuration. A/B/C question-weighted means are 70.9/79.9/89.6, pass rates are 87.57\%/98.87\%/100.00\%, and fractions scoring $\ge80$ are 7.3\%/58.8\%/99.4\%. Counts below the score threshold are 22/2/0.}
\label{fig:scorehist}
\end{figure}

Figure~\ref{fig:scorehist} shows the distributions shifting toward higher scores. Question-weighted means are 70.9, 79.9, and 89.6 for A, B, and C. At a pass threshold of 60, pass rates rise from 87.57\% to 98.87\% to 100.00\%. The fractions scoring at least 80 are 7.3\%, 58.8\%, and 99.4\%. C reduces low-scoring answers and brings nearly all answers above 80.

\begin{figure}[htbp]
\centering
\includegraphics[width=\linewidth]{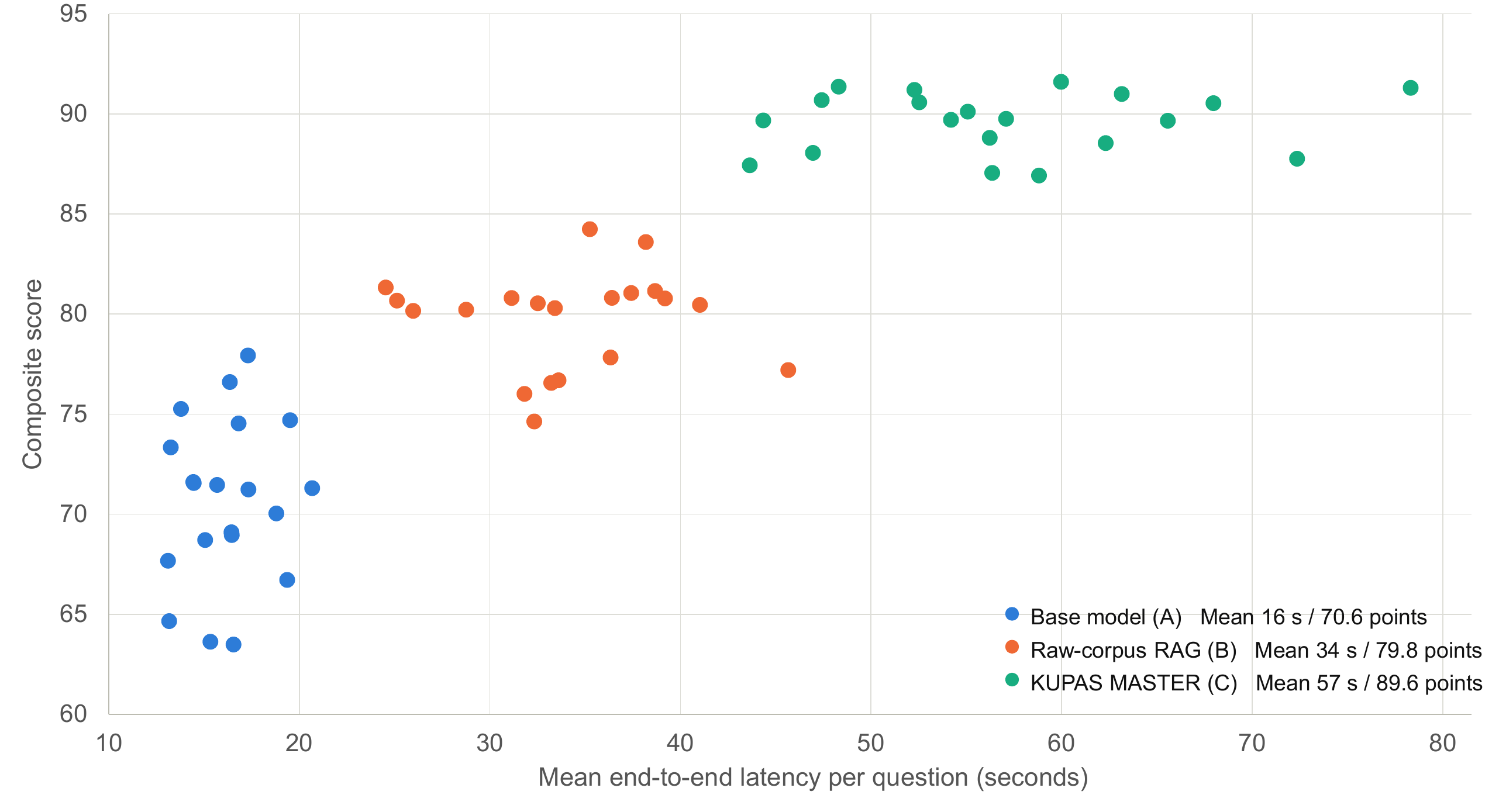}
\caption{Latency and score, with one point per practitioner and configuration. Mean A/B/C times are 16.2/34.0/57.1 s, rounded to 16/34/57 s in the legend. Ratios of mean times are $C/A=3.53$ and $C/B=1.68$.}
\label{fig:latency}
\end{figure}

In Figure~\ref{fig:latency}, each point pairs a practitioner's composite score with mean end-to-end time for one configuration. Equal weighting across practitioners gives A/B/C scores of 70.6/79.8/89.6 and times of 16.2/34.0/57.1 seconds; the legend rounds time to 16/34/57 seconds. Ratios calculated from unrounded means give C/A of 3.53 and C/B of 1.68. C achieves a score of 89.6 at a mean time of 57.1 seconds, delivering high-quality professional analysis within a practical response window.

\begin{figure}[htbp]
\centering
\includegraphics[width=\linewidth]{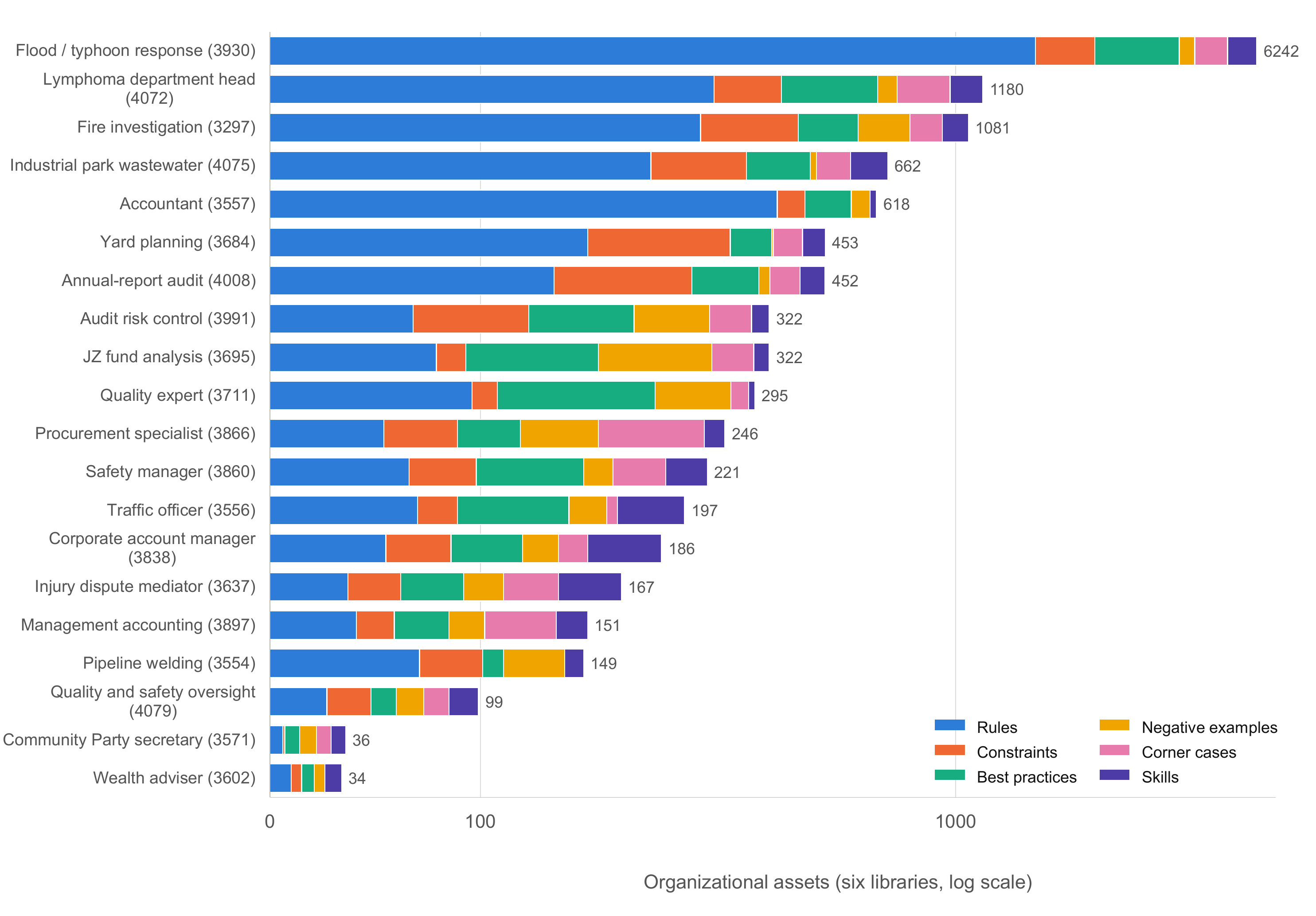}
\caption{The six libraries contain 13,113 assets: 3,550 rules, 2,892 best practices, 1,955 skills, 1,948 corner cases, 1,623 constraints, and 1,145 negative examples. All six types are present for 17/20 practitioners. The cumulative horizontal positions use a linear scale up to 200 assets and a logarithmic scale above 200; segment widths therefore do not represent type proportions.}
\label{fig:sixlibs}
\end{figure}

Figure~\ref{fig:sixlibs} counts 13,113 organizational assets across the 20 practitioners. Rules and best practices account for 49.1\%; the remaining assets supply skills, constraints, failure experience, and unusual situations. Counts and type distributions vary substantially. Seventeen practitioners have all six types, showing how the same structure accommodates different professional settings.

\begin{figure}[htbp]
\centering
\includegraphics[width=\linewidth]{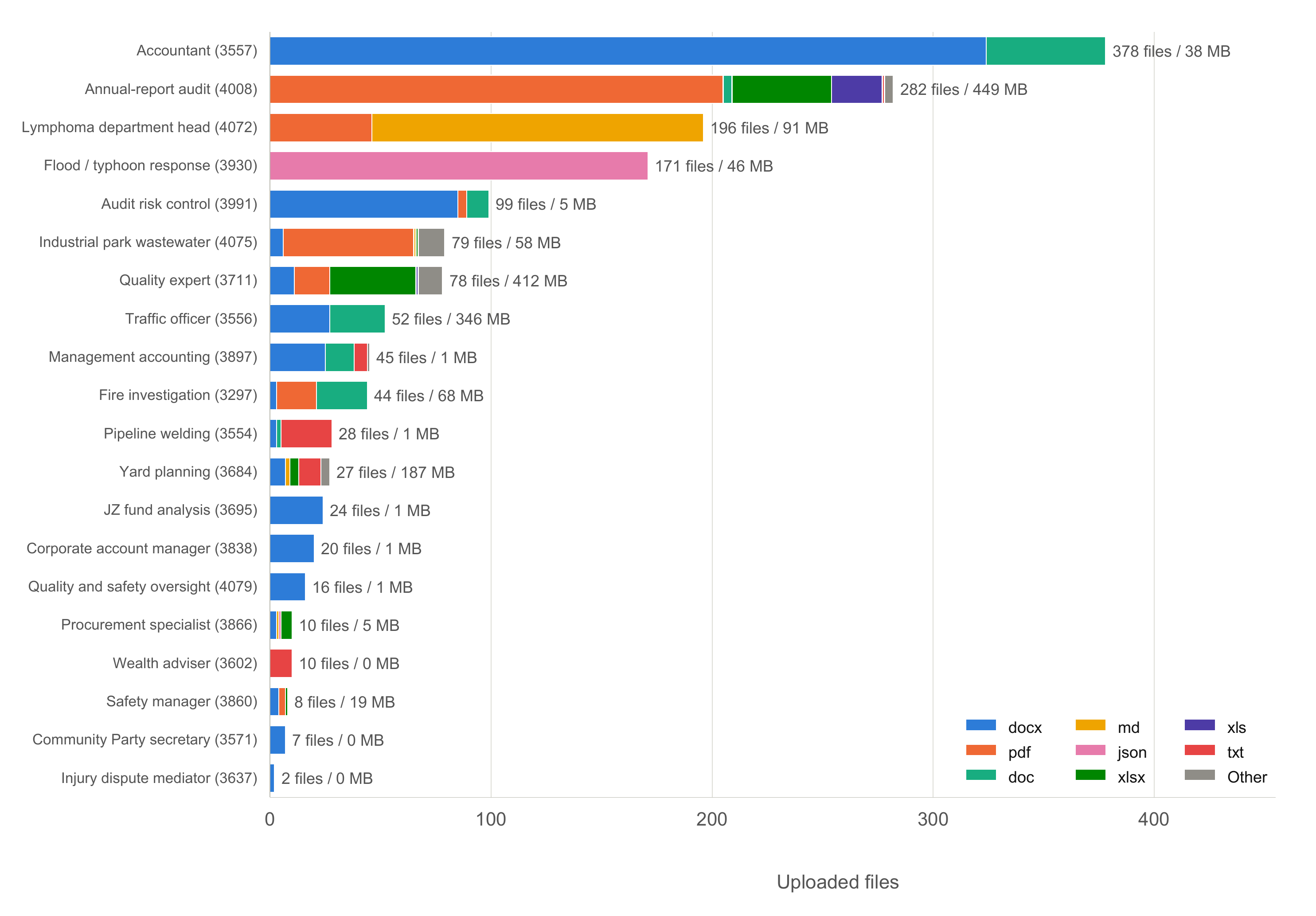}
\caption{The uploaded corpus contains 1,576 files, 1.73 GB, and 13 formats. Individual file counts range from 2 to 378, covering documents, spreadsheets, structured data, images, and recordings. MB labels are rounded to whole decimal megabytes; 0 MB denotes less than 0.5 MB, not an empty corpus.}
\label{fig:corpusfiles}
\end{figure}

Figure~\ref{fig:corpusfiles} shows 1,576 source files in 13 formats, with 2 to 378 files per practitioner. Formats and storage sizes vary widely. The platform organizes documents, tables, structured data, and multimedia while preserving source relationships for extraction and consolidation.

\begin{figure}[htbp]
\centering
\includegraphics[width=\linewidth]{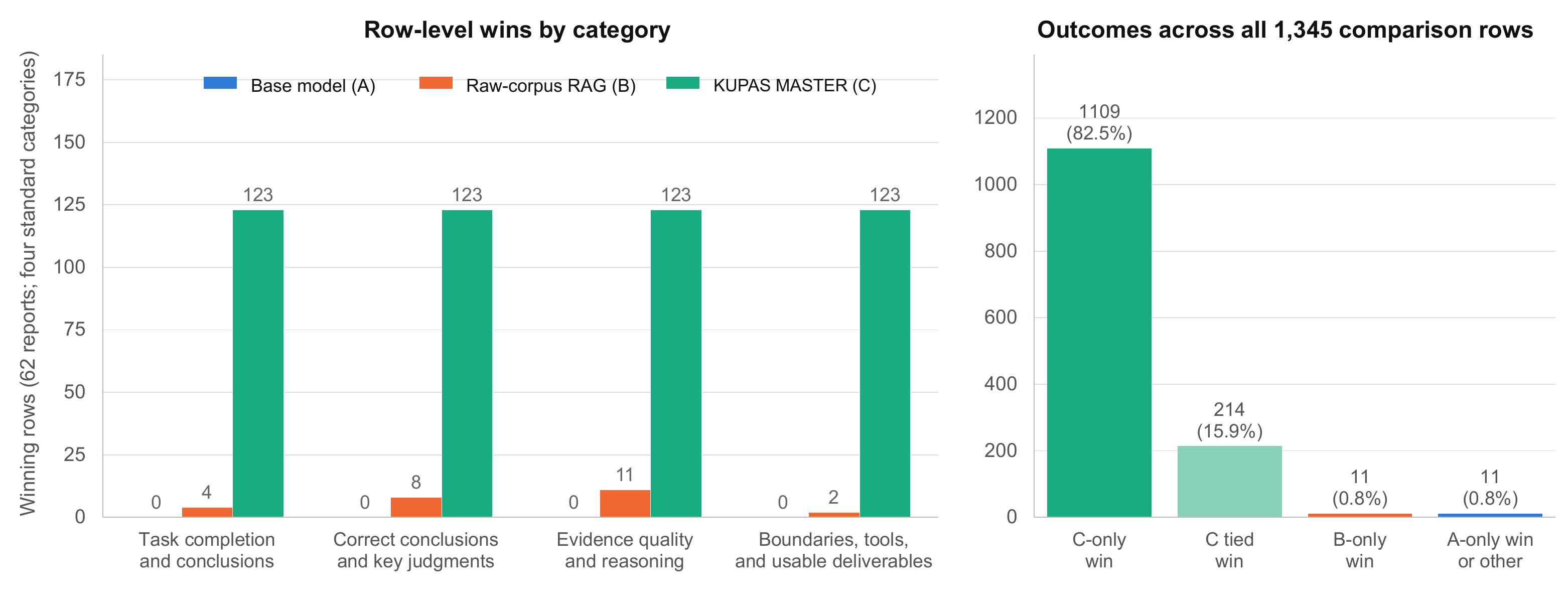}
\caption{Judgments in three-configuration comparison reports. Left: wins including ties across 123 rows per category in 62 reports. Right: outcomes across 217 retained reports and 1,345 rows, with 1,109 C-only wins (82.5\%) and 214 C ties (15.9\%).}
\label{fig:qawinners}
\end{figure}

The left panel of Figure~\ref{fig:qawinners} uses 62 reports with four standard judgment categories. Each category has 123 comparison rows, and C wins every row. Wins include ties, so a row can count for more than one configuration. The right panel uses all 217 retained reports and 1,345 comparison rows: C wins alone on 1,109 rows (82.5\%) and ties on 214 (15.9\%); B wins alone on 11 (0.8\%), and A-only wins or other outcomes account for 11 (0.8\%). C calls skills in 166 reports and generates files in 116. These report-level comparisons supplement the 177-question scored evaluation.

\begin{figure}[htbp]
\centering
\includegraphics[width=\linewidth]{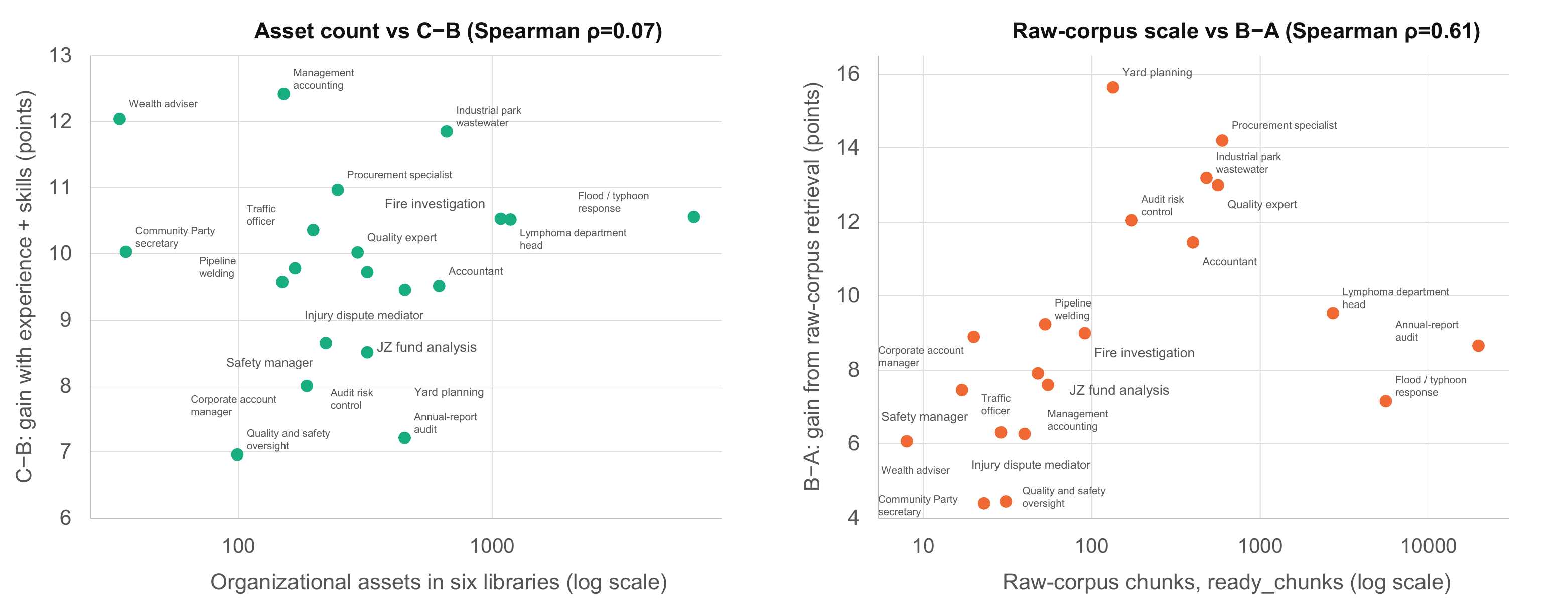}
\caption{Scale and score gains ($n=20$). Left: the Spearman correlation between organizational asset count and $C-B$ is 0.07. Right: the correlation between raw chunk count and $B-A$ is 0.61.}
\label{fig:scale}
\end{figure}

Figure~\ref{fig:scale} shows a weak relationship between organizational asset count and the C-minus-B gain across 20 practitioners, with Spearman correlation 0.07. Raw chunk count has a positive correlation of 0.61 with the B-minus-A gain. The relationship between scale and score gain therefore differs between stages, with experience organization and use adding value beyond raw retrieval.

\begin{figure}[htbp]
\centering
\includegraphics[width=\linewidth]{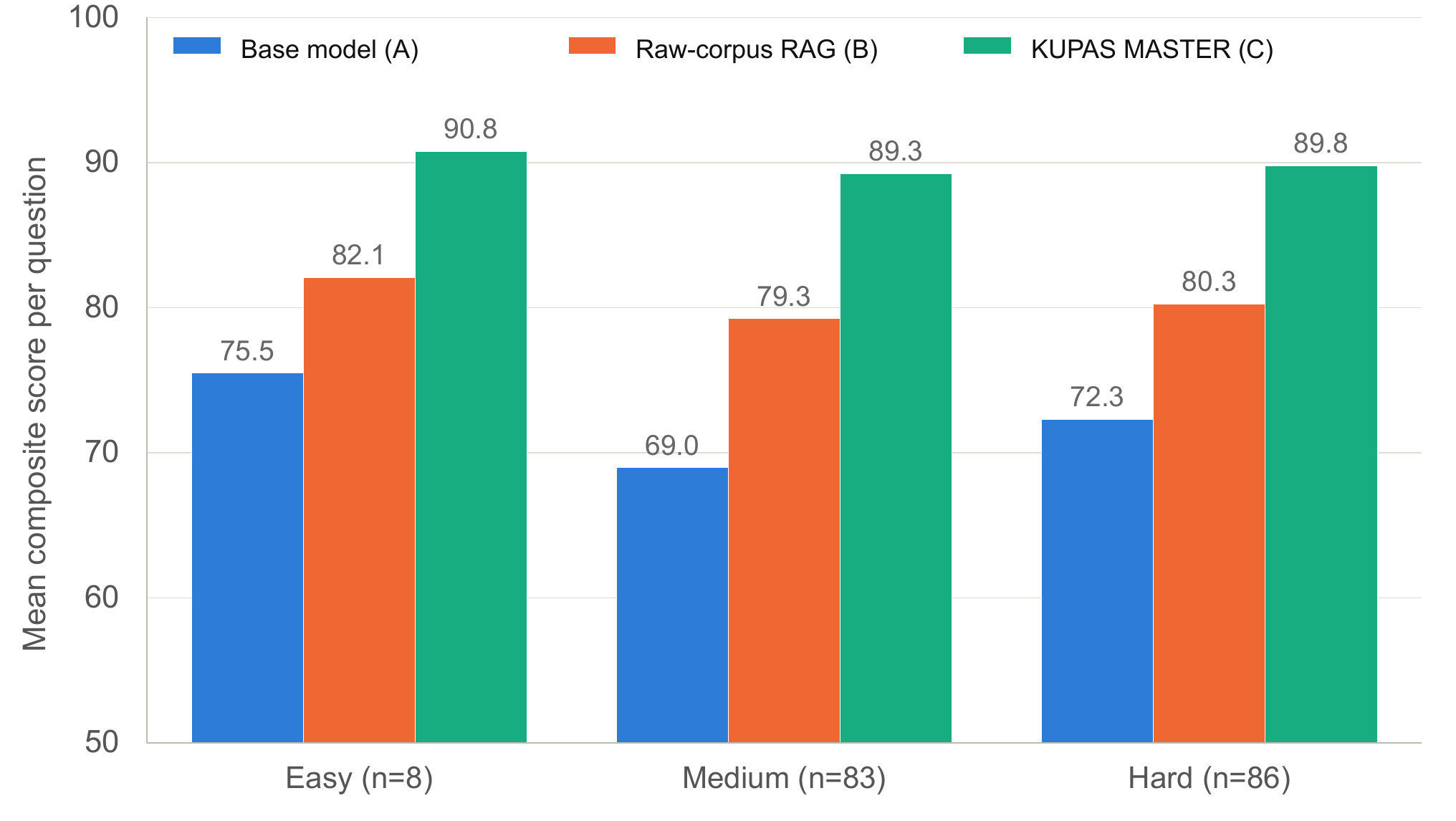}
\caption{Question-weighted composite scores by difficulty. A/B/C means are 75.5/82.1/90.8 for easy questions ($n=8$), 69.0/79.3/89.3 for medium questions ($n=83$), and 72.3/80.3/89.8 for hard questions ($n=86$). C exceeds both baselines and scores above 89 in all groups.}
\label{fig:difficulty}
\end{figure}

Figure~\ref{fig:difficulty} groups the questions by their recorded difficulty labels: 8 easy, 83 medium, and 86 hard. C's mean scores are 90.8, 89.3, and 89.8, above both baselines in every group. Its gains over B are about 8.7, 10.0, and 9.5 points. The figure reports the size of each group, including 8 easy questions.

\end{document}